\documentclass[final,5p,times,twocolumn]{elsarticle}

\usepackage{amsmath,amssymb,amsfonts,bm}
\DeclareMathOperator*{\argmax}{argmax}
\DeclareMathOperator*{\argmin}{argmin}

\usepackage{graphicx}
\usepackage{textcomp}
\usepackage{xcolor}
\usepackage{booktabs}
\usepackage{balance}

\usepackage{siunitx}
\DeclareSIUnit\rpm{rpm}

\usepackage[super]{nth}
\usepackage[export]{adjustbox}

\usepackage{caption}
\usepackage{subcaption}

\usepackage{enumitem}
\setlist{leftmargin=15pt}

\usepackage [autostyle, english = american]{csquotes}
\MakeOuterQuote{"}

\usepackage{array}
\newcommand{\PreserveBackslash}[1]{\let\temp=\\#1\let\\=\temp}
\newcolumntype{C}[1]{>{\PreserveBackslash\centering}p{#1}}
\newcolumntype{R}[1]{>{\PreserveBackslash\raggedleft}p{#1}}
\newcolumntype{L}[1]{>{\PreserveBackslash\raggedright}p{#1}}

\newcommand{\tablehdg}[1]{\textbf{#1}}
\newcommand{\tablesize}{\scriptsize}

\usepackage{algorithm}
\usepackage[noend]{algpseudocode}
\makeatletter
\def\BState{\State\hskip-\ALG@thistlm}
\makeatother

\def\BibTeX{{\rm B\kern-.05em{\sc i\kern-.025em b}\kern-.08em
    T\kern-.1667em\lower.7ex\hbox{E}\kern-.125emX}}

\usepackage{hyperref} 
\hypersetup{
  colorlinks,
  citecolor=black,
  filecolor=black,
  linkcolor=black,
  urlcolor=black,
  pdfauthor={},
  pdfsubject={},
  pdftitle={}
}
\usepackage{cleveref} 

\def\equationautorefname~#1\null{(#1)\null}

\usepackage{scalerel}
\usepackage{tikz}
\usetikzlibrary{svg.path}
\definecolor{orcidlogocol}{HTML}{A6CE39}

\DeclareRobustCommand\orcidicon[1]{\href{https://orcid.org/#1}{\mbox{\scalerel*{
        \begin{tikzpicture}[yscale=-1,transform shape,scale=0.08]
        \fill[orcidlogocol] svg{M256,128c0,70.7-57.3,128-128,128C57.3,256,0,198.7,0,128C0,57.3,57.3,0,128,0C198.7,0,256,57.3,256,128z};
        \fill[white] svg{M86.3,186.2H70.9V79.1h15.4v48.4V186.2z}
        svg{M108.9,79.1h41.6c39.6,0,57,28.3,57,53.6c0,27.5-21.5,53.6-56.8,53.6h-41.8V79.1z M124.3,172.4h24.5c34.9,0,42.9-26.5,42.9-39.7c0-21.5-13.7-39.7-43.7-39.7h-23.7V172.4z}
        svg{M88.7,56.8c0,5.5-4.5,10.1-10.1,10.1c-5.6,0-10.1-4.6-10.1-10.1c0-5.6,4.5-10.1,10.1-10.1C84.2,46.7,88.7,51.3,88.7,56.8z};
        \end{tikzpicture}
}{|}}}}

\definecolor{dark_red}{rgb}{0.4, 0.0, 0.0}
\definecolor{light_red}{rgb}{0.8, 0.1, 0.1}
\definecolor{dark_green}{rgb}{0.0, 0.4, 0.0}
\definecolor{light_green}{rgb}{0.1, 0.8, 0.1}
\definecolor{dark_blue}{rgb}{0.0, 0.0, 0.4}
\definecolor{light_blue}{rgb}{0.1, 0.1, 0.8}
\definecolor{dark_violet}{rgb}{0.4, 0.1, 0.4}
\definecolor{light_violet}{rgb}{0.8, 0.1, 0.8}
\definecolor{dark_orange}{rgb}{0.4, 0.4, 0.1}
\definecolor{light_orange}{rgb}{0.8, 0.6, 0.1}
\definecolor{tab_blue}{RGB}{31, 119, 180}
\definecolor{tab_orange}{RGB}{255, 127, 14}
\definecolor{tab_green}{RGB}{44, 160, 44}
\definecolor{tab_red}{RGB}{214, 39, 40}

\newcommand{\tvec}{\xi}
\newcommand{\massquared}{w_i^2}
\newcommand{\tij}{t_{ij}}

\usepackage[printonlyused]{acronym}
\acrodef{uav}[UAV]{Unmanned Aerial Vehicle}
\acrodef{tcn}[TCN]{Temporal Convolutional Network}
\acrodef{imu}[IMU]{Inertial Measurement Unit}
\acrodef{ekf}[EKF]{Extended Kalman Filter}
\acrodef{eskf}[ESKF]{Error State Kalman filter}
\acrodef{gnss}[GNSS]{Global Navigation Satellite System}
\acrodef{sar}[SAR]{Search and Rescue}
\acrodef{darpa}[DARPA]{Defense Advanced Research Projects Agency}
\acrodef{subt}[SubT]{Subterranean Challenge}
\acrodef{lo}[LO]{LiDAR Odometry}
\acrodef{vo}[VO]{Visual Odometry}
\acrodef{lio}[LIO]{LiDAR-Inertial Odometry}
\acrodef{vio}[VIO]{Visual-Inertial Odometry}
\acrodef{lidar}[LiDAR]{Light Detection And Ranging}
\acrodef{dof}[DOF]{Degree of Freedom}
\acrodef{mems}[MEMS]{Micro-Electro-Mechanical System}
\acrodef{rlg}[RLG]{Ring Laser Gyroscope}
\acrodef{mas}[MAS]{Motor Angular Speed}
\acrodef{esc}[ESC]{Electronic Speed Controller}
\acrodef{rpm}[RPM]{Revolutions per Minute}
\acrodef{pwm}[PWM]{Pulse Width Modulation}
\acrodef{fcu}[FCU]{Flight Controller Unit}
\acrodef{cog}[COG]{Center of Gravity}
\acrodef{map}[MAP]{Maximum a Posteriori}
\acrodef{lm}[LM]{Levenberg-Marquardt}
\acrodef{slam}[SLAM]{Simultaneous Localization and Mapping}
\acrodef{sfm}[SFM]{Structure From Motion}
\acrodef{maslo}[MAS-LO]{Motor Angular Speed LiDAR Odometry}
\acrodef{liosam}[LIO-SAM]{LiDAR-Inertial Odometry Smoothing and Mapping}
\acrodef{rtk}[RTK]{Real-time Kinematic}
\acrodef{ape}[APE]{Absolute Pose Error}
\acrodef{ate}[ATE]{Absolute Translation Error}
\acrodef{ave}[AVE]{Absolute Velocity Error}
\acrodef{are}[ARE]{Absolute Rotation Error}
\acrodef{isam2}[iSAM2]{Incremental Smoothing and Mapping}
\acrodef{lm}[LM]{Levenberg-Marquardt}
\acrodef{indi}[INDI]{Incremental Nonlinear Dynamic Inversiond}
\acrodef{ukf}[UKF]{Unscented Kalman Filter}
\acrodef{ikdtree}[ikd-Tree]{Incremental K-Dimensional Tree}

\journal{Robotics and Autonomous Systems}

\begin{document}

\begin{frontmatter}

\title{Motor Angular Speed Preintegration for Multirotor UAV State Estimation}
  \tnotetext[t1]{This work was funded by CTU grant no SGS23/177/OHK3/3T/13, by the Czech Science Foundation (GAČR) under research project no. 23-06162M and by the European Union under the project Robotics and advanced industrial production (reg. no. CZ.02.01.01/00/22\_008/0004590).}

  \author{Mat\v{e}j Petrl\'{i}k\corref{cor1}} 
  \ead{matej.petrlik@fel.cvut.cz}
  \author{Filip Nov\'{a}k}
  \ead{filip.novak@fel.cvut.cz}
  \author{Robert P\v{e}ni\v{c}ka}
  \ead{robert.penicka@fel.cvut.cz}
  \author{Martin Saska}
  \ead{martin.saska@fel.cvut.cz}

  \cortext[cor1]{Corresponding author.}

  \affiliation{organization={Department of Cybernetics, Faculty of Electrical Engineering, Czech Technical University in Prague},
            postcode={166 36}, 
            city={Prague 6},
            country={Czech Republic},
            }

\begin{abstract}
  A precise state estimate is crucial for a tight feedback control that enables agile and near-obstacle flights of \acsp{uav}.
  The state-of-the-art methods fuse slow pose measurements with high-frequency inertial measurements to obtain a precise state estimate.
  However, the inertial measurements from the \acs{imu} onboard the \acs{uav} are degraded by vibrations from spinning propellers and the precision of the estimated state suffers.
  We propose a novel approach based on the preintegration of accelerations obtained from motor speeds.
  We show that the accelerations obtained in this manner can be used for state propagation on their own to achieve better precision without including the \acs{imu}.
  Further, we propose a factor composed of the preintegrated motor speeds that can be directly employed in factor graph optimization frameworks.
  We combine our factor with \acs{lidar} measurements into the proposed \acl{maslo} (\acs{maslo}) algorithm for precise state estimation, which we open-source.
  Lastly, we evaluate the estimation precision against a state-of-the-art inertial algorithm \acs{liosam} to show 28\% improvement in position and 65\% in velocity estimation accuracy, 14\% lower measurement lag, and high robustness to wrong parameter values.
\end{abstract}




\end{frontmatter}

 \section*{Supplementary Materials}
 \label{sec:supplemenatary_materials}
 The paper is supported by the materials available at \href{http://mrs.felk.cvut.cz/papers/petrlik2026maslo}{mrs.felk.cvut.cz/papers/petrlik2026maslo}.

\section{Introduction}
\label{section:introduction}

In this paper, we propose an \ac{uav} state estimation method based on the angular speed measurements of each motor of a multi-rotor \ac{uav}.
Localization methods based on \ac{lidar} scans and/or camera images often rely on inertial measurements to increase their robustness.
\ac{lidar}-based methods benefit from inertial measurements especially during aggressive motions \cite{lee2024lidar} and visual methods can rely on \ac{imu} during a shortage of visual features \cite{sahili2023SurveyVisual}. 
Furthermore, thanks to inertial measurements, methods based on either sensing modality can provide state estimation even between low-rate sensor measurements.
However, although these inertial methods achieve impressive results on hand-carried and ground-vehicle-mounted datasets, the performance drops as measurements become degraded by the vibrations of the \ac{uav}'s propellers when deployed on \acp{uav}. 
Moreover, inertial methods are sensitive to the calibration of extrinsic and intrinsic parameters, and cheap \ac{mems} \acp{imu} used in sensor packages exhibit high measurement noise.
To eliminate these issues, we propose a novel method based on fusion of motor speeds, which are measured precisely by the \acp{esc} driving the motors and are thus vibration-free.

\begin{figure}[t!]
  \centering
  \adjincludegraphics[width=1.0\linewidth, trim={{0.05\width} {0.04\height} {0.04\width} {0.10\height}}, clip]{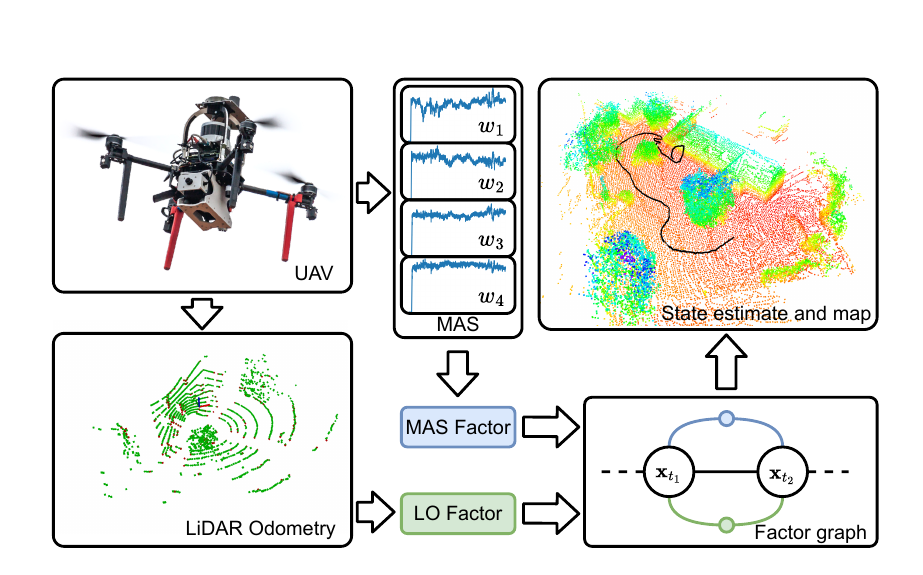}
  \caption{
    The motor angular speeds of individual motors are preintegrated into a \acs{mas} factor, which is combined with a \acs{lidar} odometry factor in a factor graph. Thereafter, an estimate of the \acs{uav} state and a map in the form of downsampled \acs{lidar} scans registered into a global coordinate frame is obtained.
  }
  \label{fig:intro_img}
\end{figure}

The majority of \ac{uav} applications are nowadays conducted outdoors using \ac{gnss} localization with a safe distance from obstacles.
In contrast, indoor flights, which are more and more demanded by the industry, situate the \ac{uav} in a constrained environment with a high risk of collision, thereby requiring precise localization and state estimation using onboard sensors.
The indoor deployment of \acp{uav} holds perhaps the greatest potential benefit for humanity in the form of \ac{sar} operations in building wreckage \cite{kratky2021exploration}, burning structures \cite{walter2022fr}, areas with nuclear radiation \cite{stibinger2020ral}, unstable mine tunnels \cite{petrlik2020robust}, or collapsed caves \cite{petracek2021caves}.
As these situations require swift action to prevent fatal injuries, robots (\acp{uav} in particular) present great potential to speed up rescue operations by providing situational awareness for human rescuers.
The \ac{darpa} identified this point as a potentially disruptive technology and, in order to promote accelerated research and provide funding, organized the \ac{darpa} \ac{subt} \cite{orekhov2023inspiring}, where state estimation was a critical component necessary for success \cite{ebadi2023present}.
Thanks to the improved precision of the proposed method, the \acp{uav} flying indoor become safer and can reach tighter areas with lower risk of collision with obstacles.

Another case where high-precision state estimation is necessary is in agile flying~\cite{foehn2022agilicious,penicka2022minimum}. 
While flying at velocities near the hover state, an imprecise state estimate is often sufficient for a stable flight.
However, during highly dynamic flights, the control inputs must be calculated based on accurate feedback to assure stable flight without oscillations.
So far, agile-flight experiments are performed mostly with the aid of motion-capture localization systems with millimeter precision \cite{romero2022time}.
Achieving such precision in aggressive maneuvers with onboard sensors is a challenge in which precise estimation of not only the pose, but, more importantly, also the velocity and acceleration play a crucial role.

To satisfy the need for precision, accuracy, and reliability of localization in the above-mentioned applications, we propose a novel approach to state estimation based on preintegration of angular velocities of the \ac{uav}'s motors.
The underlying principle of the method is visualized in~\autoref{fig:intro_img} and can be summarized as follows:
The \ac{mas} of each propeller measured by the \ac{esc} is passed through the propulsion model of the \ac{uav} to obtain linear and angular accelerations. 
Direct integration of the acquired acceleration would be highly inefficient, as the integration would have to be repeated every time the initial state changed.
To remove the dependency on the initial state, the preintegration collects multiple \ac{mas} measurements into a single delta measurement. 
This delta measurement can then be used to propagate any \ac{uav} state forward in time.
As such, the delta measurement can be used to predict the current \ac{uav} state without the need to repeat the integration every time a new \ac{mas} measurement arrives.
The delta measurement forms a \ac{mas} factor which is introduced into a factor graph estimation framework.
To constrain the drift from integrating accelerations the position and orientation have to be constrained by a localization source, in our case a \ac{lo} algorithm.
Finally, the full state estimate is obtained by performing fixed-lag smoothing on the factor graph.

The \ac{mas}-based estimation is a challenging task due to the complex aerodynamics of the \ac{uav} propellers.
If the aerodynamic effects are not modeled, the resultant thrusts and torques do not correspond to the actual thrusts and torques produced by the motors, causing the estimation to become biased.
Moreover, the mass distribution on the \ac{uav} is typically not centered due to limited mounting space for sensors and other payload.
The \ac{cog} is then offset from the geometrical center of the \ac{uav}, which is compensated by faster rotation of some propellers (see~\autoref{fig:noise} for rotation speeds of individual propellers of the \ac{uav} used for evaluation in this paper).
External disturbances such as wind or ground effect also temporarily influence the motor speeds of the propellers.
Lastly, the accuracy of orientation obtained solely from the \ac{mas} model suffers from the double integration from angular accelerations obtained from the model.

Existing approaches utilize motor speed feedback mostly to improve control performance and robustness to external disturbances.
Some works apply motion models based on \ac{mas} to improve performance of \ac{vio} algorithms or constrain the estimated velocities in case of visual localization failures.
However, these approaches still rely primarily on \ac{imu} measurements.
We summarize the available literature regarding \ac{mas}-based state estimation in~\autoref{section:mas_sota}.

In this paper, we show that the linear accelerations obtained from the \ac{uav} motion model driven by \ac{mas} exhibit much lower amplitude of noise compared to \ac{imu} measurements as seen in~\autoref{fig:noise_compare}.
We show that the \ac{mas} factor can replace the \ac{imu} factor in inertial methods and thus eliminate the problems with \ac{imu}.
By introducing the \ac{mas} factor in a state-of-the-art localization method we have achieved decrease in the position~(28\%) and linear velocity~(65\%) absolute errors at the cost of an increase in orientation error~(20\%) compared to the same method using inertial measurements (\autoref{tab:evaluation}).
The empirical analysis further demonstrates that the linear accelerations obtained from \ac{mas} model have lower measurement lag, which is an important property for feedback control in agile flight.
Our parameter sensitivity study also reveals that the \ac{mas}-based state estimation method is robust to wrong choice of parameters and even wrong physical measurements of the \ac{uav} dimensions.
Another advantage of the proposed method is directly available angular accelerations that are crucial for snap tracking in \ac{indi} controllers~\cite{tal2020accurate,sieberling2010robust}.
Without our approach, the angular accelerations have to be either numerically differentiated from angular rates measured by \ac{imu}~\cite{tal2020accurate} or predicted in a predictive filter~\cite{sieberling2010robust}.

\subsection{Contributions}
\label{sec:contributions}

A new method for state estimation based on fusion of \ac{mas} measurements is proposed in this paper.
We show that building the estimation framework using a \ac{mas}-based model instead of inertial measurements provides less noisy linear acceleration observations, which allows for more accurate position and velocity estimation.
Furthermore, we show that \ac{mas} measurements are sufficient on their own for the propagation of \ac{uav} state and thus can be employed instead of inertial measurements in state estimation methods.
Based on these findings, we design a novel \ac{mas} preintegration scheme operating on the tangent space of the \ac{uav} state manifold.
The preintegration happens in a local coordinate space and is thus invariant to the initial state. 
Therefore, the preintegration does not have to be recomputed after every linearization step of the non-linear optimization methods, which is a key property for real-time state estimation.
We facilitate the extension of factor-graph-based inertial localization methods by providing a \ac{mas} factor obtained from the preintegrated measurements, which can directly replace the inertial factor.
Next, we incorporate the \ac{mas} preintegration, \ac{mas} factor, and \ac{lo} method into a novel state estimation algorithm \textit{\ac{maslo}} to showcase the feasibility of \ac{mas}-based estimation.
We opensource our work and provide the \ac{mas} preintegration and \ac{mas} factor\footnote{\url{https://github.com/ctu-mrs/mas_factor}}, as well as the entire localization method \ac{maslo}\footnote{\url{https://github.com/ctu-mrs/maslo}} for the \ac{uav} community to encourage further development of \ac{mas}-based methods.
As datasets for evaluation of the proposed method were not available, we recorded six datasets with various motions, \ac{mas} measurements, \ac{imu} measurements, and \ac{rtk} ground truth for evaluation of the proposed algorithm. 
We provide these datasets\footnote{\url{https://github.com/ctu-mrs/mas_datasets}} for the future evaluation of new \ac{mas}-based state estimators.
We show a comprehensive evaluation of our approach, including comparison with state-of-the-art \textit{\ac{liosam}} based on multiple metrics, analysis of sensitivity to parameter values, and measurement lag evaluation.
In comparison with \ac{liosam}, \ac{maslo} reaches 28\% lower \ac{ate} and 65\% lower \ac{ave} at the cost of slightly higher 20\% \ac{are} on average over all datasets.


\subsection{Paper organization}
\label{section:organization}
This paper is organized as follows.
A survey of the literature on inertial localization methods, their limitations when deployed on \acp{uav}, and existing approaches that incorporate \ac{mas} measurements into state estimation and control is provided in \autoref{section:related_work}.
A high-level description of the proposed \ac{mas}-based state estimation approach is given in \autoref{section:proposed_solution}.
The notation and mathematical preliminaries are introduced in \autoref{section:preliminaries}.
The \ac{uav} dynamics model that maps \ac{mas} measurements to linear and angular accelerations, together with the motion equations and noise model, is presented in \autoref{section:mas_model}.
The preintegration of \ac{mas} measurements in the tangent space of the \ac{uav} state manifold is derived in \autoref{section:mas_preintegration}.
The \ac{mas} binary factor and the \ac{mas} bias binary factor are described in \autoref{section:mas_factors}.
The complete \ac{maslo} localization algorithm, which combines the \ac{mas} factor with a \ac{lidar} odometry factor, is presented in \autoref{section:maslo}.
A comprehensive experimental evaluation, including noise analysis, runtime profiling, measurement lag comparison, parameter sensitivity study, and trajectory error evaluation against \ac{rtk} ground truth, is provided in \autoref{section:evaluation}.
Conclusions and directions for future work are given in \autoref{section:conclusion}.


\section{Related Work}
\label{section:related_work}

To achieve precise state estimation in real-world flights outside of motion capture systems, onboard sensors are employed in localization methods, with \ac{lo} and \ac{vo} being the two most prominent approaches, thanks to precise 3D environment scans of \acs{lidar} sensors and the wide availability of inexpensive cameras.
The estimation precision, especially during highly dynamic motions, can be further enhanced by including inertial measurements from \ac{imu} into \ac{lo} or \ac{vo}.
These \acf{lio} methods~\cite{zhang2017low,qin2020lins,xu2021fast,xu2022fast,he2023point,bosse2012zebedee,shan2020lio,geneva2018lips} surveyed in~\cite{lee2024lidar} and \acf{vio} methods~\cite{forster2016svo,leutenegger2015keyframe,qin2018vins,geneva2020openvins,rosinol2020kimera,lupton2011visual} reviewed in~\cite{sahili2023SurveyVisual} provide a state estimate at the \ac{imu} data rate, which is often a magnitude higher than the data rate of the primary sensors. 
This allows for more frequent action inputs to be applied by the feedback controller, ensuring a smooth flight with low control error.
In this section, we first introduce the inertial methods, then we discuss the limitations and potential issues with using inertial measurements in state estimation.
Due to these limitations we propose to build the state estimation on \ac{mas} instead.
Finally we review the state-of-the-art methods that utilize \ac{mas} for state estimation and for improving control algorithms and we also note how our approach differs from these methods.

\subsection{Inertial methods}
\label{section:inertial_methods}

Most of the latest \ac{lio} methods are tightly coupled~\cite{lee2024lidar}, meaning that the inertial measurements are processed together with the \acs{lidar} measurements in a unified framework, which fully exploits the complementary properties of both sensing modalities.
In contrast, loosely coupled methods process \ac{imu} and \acs{lidar} measurements individually, thereafter fusing the results.
For instance, in loosely-coupled LOAM~\cite{zhang2017low}, the \ac{lidar} scan that is being aligned to a previous one is firstly rotated using the orientation of the sensor estimated by integrating the inertial measurements.
Additionally, the motion distortion of a scan from a rotating \acs{lidar} on a moving platform is compensated for by the velocity, which is also estimated using the \ac{imu}.
Tightly-coupled methods, in general, achieve higher estimation accuracy~\cite{lee2024lidar}, which was confirmed in the proposed approach that is also tightly-coupled.

The frameworks utilized in tightly coupled methods can be divided into filter-based~\cite{qin2020lins,xu2021fast,xu2022fast,he2023point,geneva2020openvins} and graph-optimization-based~\cite{bosse2012zebedee,mur2015orb,mur2017orb,forster2016svo,leutenegger2015keyframe,qin2018vins,shan2020lio,rosinol2020kimera} approaches.
Filtering approaches offer fast computation times, even with complex models that fuse multi-modal sensor measurements.
However, their ability to handle delayed measurements is limited, as they typically integrate only the latest measurements.
Most filter-based methods rely on the classical Kalman filter and its advanced implementations, such as the iterated \ac{eskf} in LINS~\cite{qin2020lins}, to provide precise localization with low computational complexity.
Fast-LIO~\cite{xu2021fast} further improves the computation speed by introducing a new formula to compute the Kalman gain.
The computation complexity depends on the state dimension instead of the measurement dimension, which allows for the fast fusion of thousands of feature points.
Fast-LIO2~\cite{xu2022fast} builds upon~\cite{xu2021fast} and improves accuracy by eliminating the need for feature extraction.
The fusion of raw points was made possible with the further increase of speed provided by employing \ac{ikdtree} for efficient downsampling and nearest neighbor search.
Point-LIO~\cite{he2023point} goes even further by fusing each point individually as it arrives, achieving output odometry frequency in the order of kilohertz.

Optimization-based methods build a graph from poses and measurements as vertices and edges, respectively, thereby representing relative constraints between the poses.
The graph represents a history of measurements (typically of a fixed time horizon to keep the computation time-bounded) that allows for fusing delayed measurements.
The optimization problem is then defined as the minimization of the sum of the least squares errors of all constraints.
Zebedee~\cite{bosse2012zebedee} is an early example of an optimization-based approach that combined a spring-mounted 2D \ac{lidar} with \ac{imu} to estimate a full 6-\ac{dof} trajectory.
Among open-source optimization frameworks~\cite{juric2021ComparisonGraph} capable of solving nonlinear optimization problems defined as graphs, the g\textsuperscript{2}o~\cite{kummerle2011g2o} stands out.
This framework is used in ORB-SLAM~\cite{mur2015orb, mur2017orb} for pose estimation and in SVO~\cite{forster2016svo} for visual odometry. 
Across different domains, Ceres~\cite{agarwal2012ceres} is the most widely used thanks to its high-quality solutions with low computation time.
Most notably, it is used to optimize the nonlinear optimization problems in OKVIS~\cite{leutenegger2015keyframe}, VINS~\cite{qin2018vins}, and OpenVINS~\cite{geneva2020openvins}.
LOAM~\cite{zhang2017low} employs Ceres for optimization of the transformation between two feature point clouds.
GTSAM~\cite{dellaert2012factor} is based on defining problems as factor graphs (a generalization of pose graphs), which allows for the abstraction of many typical robotic problems, including complex state estimation problems.
This framework includes a collection of ready-to-use factors and, more importantly, allows for the easy creation of custom factors, therefore we build our approach using the GTSAM framework.
Some examples of methods running on GTSAM are LIO-SAM~\cite{shan2020lio}, where a sliding-window approach marginalizing old \ac{lidar} scans to adhere to the real-time constraints was introduced, and Kimera~\cite{rosinol2020kimera}, which estimates a 3D mesh representation of the environment in addition to \ac{vio}.
Optimization methods generally achieve higher accuracy and robustness to disturbances compared to filter-based approaches, which is why we also represent the state estimation as a sliding-window nonlinear optimization problem in the form of factor graph.
Thanks to the high accuracy, we base the implementation of the \ac{lo} part of \ac{maslo} on the LIO-SAM~\cite{shan2020lio} method. 

To obtain motion priors from \ac{imu} for fusion in both filter-based and graph-based methods, the inertial measurements are first integrated to obtain a relative pose and linear velocity.
The frequent integration of inertial measurements from high-frequency \ac{imu} leads to significant computational load and, more importantly, requires the initial pose of the integration to be known at the integration time.
The authors of~\cite{lupton2011visual} devised a preintegration approach not dependent on initial conditions, thus preintegrating \ac{imu} measurements independently of the initial state and measurements from other sensors.
Preintegration methods~\cite{lupton2011visual,forster2016manifold} alleviate the high load by integrating multiple \ac{imu} measurements into a single factor that is added to the graph when a new \ac{lidar} measurement becomes available.
Our approach employs a similar principle of preintegration to achieve independence of the initial state and speed up computation but instead of preintegrating inertial measurements, we preintegrate accelerations obtained from \ac{mas} into an \ac{mas} factor.

Factor graphs are probabilistic graph models that can represent large-scale inference problems in robotics.
In contrast to pose graphs, two types of nodes appear in any factor graph: \emph{variables}, as the quantities we want to infer, and \emph{factors}, which put probabilistic constraints on the variable values. 
The graph edges always connect a factor-variable pair. Consequently, a factor becomes a function of the variables connected to by its edges.
A factor can be connected to multiple variables, and thus a factor graph is able to model more complex problems than a pose graph, where an edge always connects exactly two poses.
Every pose graph can be easily converted to a factor graph by substituting the edges of the pose graph for factors.
We use factor graphs to represent a sliding window of preintegrated \ac{mas}, \ac{lo} measurements, and bias terms.


As mentioned above, inertial algorithms (\ac{lio} and \ac{vio}) outperform their \ac{lidar}/camera-only counterparts (\ac{lo} and \ac{vo}) in state estimation output frequency, precision, and robustness.
Thanks to the wide availability and small size of inexpensive \ac{imu} units, most of the recently developed state-of-the-art localization methods integrate inertial measurements in a tightly-coupled manner to fully complement the data obtained from the main sensor. 
However, \acp{imu} come with several caveats and challenges that must be acknowledged as they can severely limit the performance, or even prevent its use altogether. 
The specific challenges are:
    \subsubsection{Propeller-induced vibrations}
   \acp{imu} mounted on the \ac{uav} body suffer from high-amplitude propeller-induced vibrations.
    The true acceleration trajectory easily disappears in such noise (see~\autoref{fig:noise_compare}), which severely degrades the performance of inertial algorithms \cite{capriglione2020experimental}.
    Mechanical decoupling of the \ac{imu} from the \ac{uav} frame using flexible elements can damp the vibrations to some extent \cite{duan2020dynamical}, but at the cost of introducing resonant frequencies to the system \cite{li2017development}, which further deviates the inertial measurements from their true values.
    In addition to mechanical decoupling, further reduction of noise can be achieved by software filtering, specifically by passing the \ac{imu} data through low-pass or notch filters. 
    However, this introduces additional delay to the system and the frequency to be filtered out changes depending on the \ac{mas}.
    In~\cite{brossard2020denoising}, deep learning was applied to filter out the noise in gyroscope data to reduce the error in attitude estimation and improve the accuracy of state-of-the-art visual localization methods.

  \subsubsection{High sensitivity to calibration}
   Inertial methods require highly accurate extrinsic camera/\ac{lidar} to \ac{imu} transform calibration, exact timestamping of sensor data, and \ac{imu} intrinsic calibration to function as intended.
   Estimation of these parameters is possible with powerful tools, such as kalibr \cite{furgale2013unified}.
    Some methods can even estimate these parameters online \cite{qin2018online}.
    Nevertheless, a slightly wrong calibration can cause a huge negative impact on the performance of the state estimation.
    Moreover, during degenerate motions (constant acceleration and angular velocities in some or all axes) that are often present in \ac{uav} flight, online calibration degrades the estimation accuracy, as the calibrated parameters become unobservable \cite{yang2020OnlineIMU}.

    \subsubsection{9-DoF IMU requirement}
     Some algorithms (LOAM \cite{zhang2017low}, LIO-SAM \cite{shan2020lio}) require 9-axis (9-\ac{dof}) \ac{imu}, which in addition to an accelerometer and gyroscope also contain a magnetometer to provide absolute orientation relative to the Earth's magnetic field.
     Sensor packages combining \ac{lidar} or camera sensors with \ac{imu} (e.g. Ouster \acp{lidar}, Livox \ac{lidar}, Intel RealSense D435i) typically provide only 6-\ac{dof} \acp{imu}.
     Flight controllers usually contain 9-\ac{dof} \ac{imu} but due to electromagnetic interference caused by propulsion motors the measurements of magnetic field are not reliable and the \ac{fcu} must be extended with an external magnetometer located further away from the sources of electromagnetic emission.

    \subsubsection{High measurement noise} 
    The cheap \ac{mems} \acp{imu} typically used in \acp{uav} have relatively high measurement noise and bias drift compared to thermally stabilized \acp{imu} with \ac{rlg} angular measurements \cite{niu2015quantitative}.


\subsection{Motor angular speed measurements}
\label{section:mas_sota}
Measurements of motor angular speeds are often used in feedback controllers to improve their performance.
For example, motor speed feedback was incorporated into the \ac{uav} control law in~\cite{gonzalez2013real} to improve the robustness to external disturbances.
The motor speeds were measured by sensing optical markers on the \ac{uav} motors.
In \cite{tal2020accurate}, optoelectronically measured motor speeds are closing the motor speed control loop for accurate snap tracking, which is achieved by \ac{indi} controller.

The amount of literature regarding state estimation using motor speed measurements is scarce, as most of the developed localization methods rely on inertial measurements.
An \ac{ekf}-based approach that uses the propulsion model was designed in~\cite{burri2015robust} to keep the linear velocity estimates bounded in case of failure and/or reinitialization of visual localization algorithms.
The model is propagated using \ac{imu} and pressure sensor measurements together with thrust force and rotor drag force, which are calculated from measured motor speeds using offline estimated parameters. 
Similarly, VIMO~\cite{nisar2019vimo} adds model-predicted motion constraints to improve the performance of VINS-Mono~\cite{qin2018vins}.
A fusion of \ac{imu} with \ac{mas} in an \ac{ukf} is shown to be able to estimate the tilt, angular, velocity, linear velocity, and model parameters in \cite{svacha2020imu}.
The authors also show that the blade flapping moment term is necessary in the model to be able to estimate the moment of inertia reliably.
While these approaches include the motor speeds in their model, inertial measurements are still required.
We show that motor speed measurements are sufficient for \ac{uav} state propagation thanks to precise measurements of motor speeds reported by the \acp{esc}.

A learning-based state estimation approach for agile drone racing is presented in~\cite{cioffi2023learned}.
Mass-normalized collective thrust obtained from measured motor speeds was fed together with gyroscope measurements into a \ac{tcn} to estimate the relative positional displacement.
A full state estimate is then obtained in an \ac{ekf} framework propagated by the \ac{imu} measurements with relative positional displacement corrections from the \ac{tcn}.
However, the approach cannot generalize to trajectories not seen during training.
In~\cite{cioffi2023hdvio}, the model based on collective thrust is combined with a learning-based component that captures unmodeled effects such as aerodynamic drag and motion disturbances using gyroscope measurements, which improves the motion estimation accuracy and is also able to generalize to unseen trajectories.
However, compared to other model-based approaches, \cite{cioffi2023hdvio} requires a GPU for real-time performance of the neural inference.
In summary, learning-based methods in motion estimation are an active research field with great results but the field is not yet mature enough for general flights.

To the best of our knowledge, only two datasets containing motor speed measurements exists so far.
However, there are yet no dataset containing motor speeds and \ac{lidar} data.
The Blackbird indoor fast-flight dataset~\cite{antonini2018blackbird} contains motor speeds at \SI{190}{\hertz} rate with \ac{imu} measurements, virtual camera image streams, and ground truth pose from a motion capture system.
The motor speeds were obtained using custom optical motor encoders.
The VID dataset~\cite{zhang2022visual} additionally contains motor currents and ground truth external forces for evaluation of external disturbance estimation.
The motor speeds are measured at \SI{1}{\kilo\hertz} using a pair of orthogonal Hall position sensors.
The absence of datasets with motor speed and \ac{lidar} measurements motivated us to gather outdoor datasets of 6 flights with different kinds of motion, and with \ac{rtk} ground truth. 
We opensource the datasets as part of this paper.

To summarize the related work, the literature describes control methods that incorporate motor speed feedback to increase reference tracking accuracy \cite{tal2020accurate} and robustify the controller against external disturbances~\cite{gonzalez2013real,burri2015robust}.
Other methods fuse motor speeds~\cite{svacha2020imu} or collective thrusts~\cite{nisar2019vimo,cioffi2023learned,cioffi2023hdvio} with \ac{imu} measurements.
However, there is no work that would use motor speed measurements without \ac{imu} altogether, which is the case for the approach that we propose.


\section{Proposed MAS-based state estimation}
\label{section:proposed_solution}

\begin{figure}[thpb]
  \centering
  \adjincludegraphics[width=1.0\linewidth, trim={{0.08\width} {0.07\height} {0.13\width} {0.09\height}}, clip]{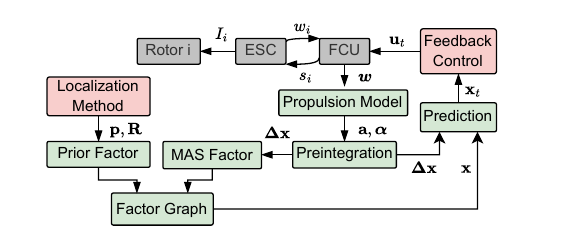}
  \caption{
    The pipeline diagram of the proposed solution shows the modules described in this paper in green. 
    Hardware parts are shown in gray. 
    Software modules that are not part of the proposed solution, but are still needed in the pipeline, are in red.
    The \acp{esc} of each of the $N$ rotors of the multirotor \ac{uav} report the \ac{mas} $w_i$ to the \ac{fcu}, which provides the vector of all \ac{mas} $\bm{w}$ to the dynamic model. 
    After passing $\bm{w}$ through the dynamic model of the \ac{uav}, the resultant linear and angular accelerations $\mathbf{a}$ and $\bm{\alpha}$ are preintegrated into the delta state $\bm{\Delta}\mathbf{x}$, which forms a full-state constraint in the form of a \ac{mas} factor.
    The position $\mathbf{p}$ and orientation $\mathbf{R}$ obtained from a localization method of choice form a prior factor that constrains the pose.
    After minimizing the residual errors in the factor graph, the full state $\mathbf{x}$ is propagated using the most recent $\bm{\Delta}\mathbf{x}$ into the current time $t$ and fed into the feedback controller, which computes the control input $\mathbf{u}$.
   Based on $\mathbf{u}$, the \ac{fcu} outputs the \ac{mas} commands $s_i$ to each \ac{esc}, which drives the rotor with current $I_i$. 
    }
  \label{fig:solution_pipeline}
\end{figure}

We propose a model-based \ac{mas} fusion approach that preintegrates the \ac{mas} measurements into a factor that can either complement inertial measurements, or replace them altogether.
A high-level description of the key components of the approach is visualized in \autoref{fig:solution_pipeline}.

The first step of the proposed approach is to obtain the \ac{mas} measurements. 
One option is to use optoelectronic sensing of reflective markers \cite{gonzalez2013real,tal2020accurate}. However, this requires additional hardware to be installed on the \ac{uav}.
The second option is using the \ac{rpm} value reported as telemetry by DShot \acp{esc}.
Most multi-rotor \acp{uav} use DShot \acp{esc} due to its lower latency compared to \ac{pwm} \acp{esc}.
Moreover, the digital transfer of commands removes the need for calibration of the pulse width.
The motor speed commands are sent from the \ac{fcu} as a throttle value in the 0--1 range to the \ac{esc} and the \ac{esc} reports back the actual \ac{rpm} value achieved by the motor over the telemetry wire.
Both optoelectronic and DShot options can be used for \ac{mas} preintegration but all presented results in this work were obtained using DShot.

The \ac{mas} measurements of all motors are propagated through the \ac{uav} propulsion model, which converts \ac{mas} measurements of individual motors into linear and angular accelerations of the \ac{cog} of the \ac{uav} where the \ac{fcu} is ideally located. 
The propulsion model depends on the number of rotors, their position relative to the \ac{cog}, and spin direction. 
The model converts \ac{mas} measurements into forces and torques acting on the \ac{cog} using the thrust and torque coefficients.
With known \ac{uav} mass and inertial matrix, the resultant force and torque are converted into linear and angular accelerations, respectively.
A detailed formal description of this model will be introduced in \autoref{section:mas_model}.

Using the raw accelerations directly in the graph optimization would be too computationally demanding.
The accelerations are obtained at the same rate as the \acp{esc} report \ac{rpm} of the motors, which is a few hundred times per second.
Adding a factor for each such measurement would cause the factor graph to grow by hundreds of nodes per second, making real-time optimization infeasible.
To allow lower-rate on-demand fusion of the obtained accelerations while preserving all available information, we employ preintegration of the \ac{mas} measurements to obtain delta values of the \ac{uav} state (as detailed in \autoref{section:mas_preintegration}), similar to how \cite{forster2016manifold} does with inertial measurements.

When a pose measurement from a localization method is available, it is converted into a prior factor that constrains the position and orientation state variables. 
At the same time, the preintegrated \ac{mas} measurements are converted into the \ac{mas} factor that constrains the relative change of the whole state vector.
To compensate for drift accumulated during the preintegration, compensate the non-zero transformation between the center of the \ac{uav} and the \ac{cog}, and to accommodate unmodeled aerodynamic effects, we add a linear and angular acceleration bias factor that integrates all of these errors into a single term.
These factors are further explained in \autoref{section:mas_factors}.

The implementation details of the proposed preintegrated \ac{mas} factor are described in \autoref{section:mas_preintegration}. 
We demonstrate the effectiveness of the \ac{mas} factor in the novel open-source localization method \ac{maslo}, which is based on the \ac{mas} measurements combined with \ac{lidar} scan matching.


\section{Notation and Preliminaries}
\label{section:preliminaries}

\begin{figure}[thpb]
  \centering
  \adjincludegraphics[width=0.5\linewidth, trim={{0.00\width} {0.00\height} {0.00\width} {0.00\height}}, clip]{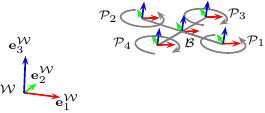}
  \caption{
    This figure shows the world frame $\mathcal{W}$ with its standard basis vectors $\mathbf{e}^\mathcal{W}_1$,  $\mathbf{e}^\mathcal{W}_2$, $\mathbf{e}^\mathcal{W}_3$.
    The \ac{uav} body frame $\mathcal{B}$ is located in the center of the \ac{uav}, and the frames of each propeller $\mathcal{P}_1$--$\mathcal{P}_4$ are rigidly attached to $\mathcal{B}$.
    The spin direction of the propellers is also visualized.
    }
  \label{fig:quadrotor_sketch}
\end{figure}

\subsection{Reference frames}
\label{section:reference_frames}
The frame in which a variable is expressed is denoted by adding the corresponding superscript to the variable.
We assume the \ac{uav} is moving in a global world frame $\mathcal{W}$, so its position is $\mathbf{p}^{\mathcal{W}}$.
The body frame $\mathcal{B}$ is fixed in the center of the \ac{uav} and is used to express, e.g., linear and angular velocities $\mathbf{v}^{\mathcal{B}}$ and $\bm{\omega}^{\mathcal{B}}$ in the \ac{uav} dynamic model.
When expressing rotations, we read $\mathbf{R}_{\mathcal{B}}^{\mathcal{W}}$ as "the rotation of the body frame in the world frame".
The last frame used is the frame of a propeller $\mathcal{P}_i$, in which the torque and drag produced by the propeller $i$ are expressed.
See \autoref{fig:quadrotor_sketch} for visualization of the individual frames.

\subsection{Special Orthogonal Group SO(3)}
\label{section:so3}

The rotation matrices in 3D form the $SO(3)$ group
\begin{equation}
  SO(3) \triangleq \left\{ \mathbf{R} \in \mathbb{R}^{3\times 3}, \mathbf{R}^{\top}\mathbf{R}=\mathbf{I}, \mathrm{det}(\mathbf{R}) = 1 \right\},
\end{equation}
with matrix multiplication as the group operation and matrix transpose as the group inverse.
The $SO(3)$ group is a smooth manifold, and $\mathfrak{so}(3)$ is its tangent space at the identity with the \textit{exponential map} operator $\mathrm{exp}([\bm{\theta}]_{\times}): \mathfrak{so}(3) \rightarrow SO(3)$:
\begin{equation}
  \mathbf{R} = \mathrm{exp}([\bm{\theta}]_{\times}) = \mathbf{I} + \frac{\mathrm{sin}(\lVert\bm{\theta}\rVert)}{\lVert\bm{\theta}\rVert}[\bm{\theta}]_{\times} + \frac{1-\mathrm{cos}(\lVert\bm{\theta}\rVert)}{\lVert\bm{\theta}\rVert}[\bm{\theta}]_{\times}^2.
\end{equation}
The inverse map is the \textit{logarithm map} $SO(3) \rightarrow \mathfrak{so}(3)$:
\begin{equation}
  \bm{\theta} = \mathrm{log}(\mathbf{R}) = \frac{\theta(\mathbf{R}-\mathbf{R}^{\top})}{2\mathrm{sin}(\theta)}
  , \quad
  \theta = \mathrm{cos}^{-1}\left(\frac{\mathrm{trace}(\mathbf{R})-1}{2}\right).
\end{equation}

We define the vector $\bm{\theta} \triangleq \mathbf{u}\theta \triangleq \bm{\omega}t \in \mathbb{R}^3$ as the integrated rotation in angle-axis form, with angle $\theta$ and unit axis $\mathbf{u}$.
The $\mathfrak{so}(3)$ space corresponds to the space of skew-symmetric matrices, which can be mapped to a corresponding vector space using the \textit{hat} $(\cdot)^{\wedge}$ and \textit{vee} $(\cdot)^{\vee}$ isomorphisms:
\begin{equation}
  \begin{gathered}
    \bm{\theta}^{\wedge} 
    = 
    \left[ 
    \begin{smallmatrix} 
    \theta_1 \\
    \theta_2 \\
    \theta_3
    \end{smallmatrix}
    \right]^{\wedge}
    =
    \left[ 
    \begin{smallmatrix} 
    0 && -\theta_3 && \theta_2 \\ 
    \theta_3 && 0 && -\theta_1 \\ 
    -\theta_2 && \theta_1 && 0
    \end{smallmatrix}
    \right]
    = 
    [\bm{\theta}]_{\times}
    \in \mathfrak{so}(3), \\
    [\bm{\theta}]_{\times}^{\vee}
    =
    (\bm{\theta}^{\wedge})^{\vee}
    =
    \bm{\theta}
    \in \mathbb{R}^{3}.
  \end{gathered}
\end{equation}

\subsection{UAV State Manifold}
\label{section:uav_state_manifold}

We define the $\textit{state}$ of the \ac{uav} as
\begin{equation}
  \mathbf{x} \triangleq \left\{\mathbf{p}^{\mathcal{W}}, \mathbf{R}^{\mathcal{W}}_{\mathcal{B}}, \mathbf{v}^{\mathcal{W}}, \bm{\omega}^{\mathcal{W}}\right\} \in \mathcal{S} \triangleq \mathbb{R}^3 \times SO(3) \times \mathbb{R}^3 \times \mathbb{R}^3,
\end{equation}
which is a smooth manifold that locally behaves as a linear space \cite{sola2018micro}.
The state consists of the position of the \ac{uav} in the world frame $\mathbf{p}^{\mathcal{W}}$, the rotation of its body frame in the world frame $\mathbf{R}^{\mathcal{W}}_{\mathcal{B}}$, its linear velocity in the world frame $\mathbf{v}^{\mathcal{W}}$, and its angular velocity in the world frame $\bm{\omega}^{\mathcal{W}}$.
The \ac{uav} state manifold is a multiplicative group, therefore it satisfies the inverse axiom
\begin{equation}
  \label{eq:inverse_axiom}
  \mathbf{x}^{-1}\mathbf{x} = \mathbf{I}.
\end{equation}
We find its tangent space by differentiating~\autoref{eq:inverse_axiom}
\begin{equation}
  \label{eq:diff_inverse_axiom}
  \mathbf{x}^{-1}\dot{\mathbf{x}} + \dot{\mathbf{x}}^{-1}\mathbf{x} = 0.
\end{equation}
Next, we rearrange~\autoref{eq:diff_inverse_axiom} to obtain the structure of the tangent space
\begin{equation}
  \label{eq:tangent_space_structure}
  \bm{\xi}^{\wedge} = \mathbf{x}^{-1}\dot{\mathbf{x}} = - \dot{\mathbf{x}}^{-1}\mathbf{x} \in \mathfrak{s} \triangleq \mathbb{R}^3 \times \mathfrak{so}(3) \times \mathbb{R}^3 \times \mathbb{R}^3,
\end{equation}
which is isomorphic to the vector space $\mathbb{R}^{12}$
\begin{equation}
  \bm{\tvec} = \left(\bm{\tvec}^{\wedge}\right)^{\vee} \triangleq \left[\mathbf{v}_{\xi},\bm{\omega}_{\xi},\mathbf{a}_{\xi},\bm{\alpha}_{\xi}\right] \in \mathbb{R}^{12}.
\end{equation}
We also need to define the \textit{retract} operator $\mathcal{R}_{\mathbf{x}_0}(\bm{\tvec}): \mathbb{R}^{12} \rightarrow \mathcal{S}$, which maps the tangent vector $\bm{\tvec}$ from the tangent space at $\mathbf{x}_0$ back to the manifold $\mathcal{S}$.
From~\autoref{eq:tangent_space_structure} we get
\begin{equation}
  \dot{\mathbf{x}} = \mathbf{x}\bm{\xi}^{\wedge},
\end{equation}
which is a differential equation with solution
\begin{multline}
  \mathbf{x}(t) = \mathcal{R}_{\mathbf{x}_0}(\bm{\tvec}) = \mathbf{x}_0\mathrm{exp}(\bm{\xi}^{\wedge}t)
  = \\
  \{\mathbf{p}_0+\mathbf{R}_0(\mathbf{v}_{\xi}t), \mathbf{R}_0\mathrm{exp}(\bm{\omega}_{\xi} t), \mathbf{v}_0 + ... \\
  \mathbf{R}_0(\mathbf{a}_{\xi}t), \bm{\omega}_0+\mathbf{R}_0(\bm{\alpha}_{\xi} t)\},
\end{multline}
  where $\mathbf{a}_{\xi}$ and $\bm{\alpha}_{\xi}$ are linear and angular accelerations respectively.

\subsection{Factor Graph}
\label{section:factor_graph}

A factor graph defines the factorization of function $f(\mathbf{X})$:
\begin{equation}
  f(\mathbf{X}) = \prod_{i=1}^{n}f_i(\mathbf{X}_i),
\end{equation}
where each factor $f_i$ depends on a subset of states $\mathbf{X}_i\subseteq\mathbf{X}=\{\mathbf{x}_1, \mathbf{x}_2, ...,\mathbf{x}_m\}$, $n$ is the total number of factors, and $m$ is the total number of states.
We are looking for the combination of states that maximizes the \ac{map} estimate:
\begin{equation}
  \label{eq:map_max}
  \mathbf{X}^{\mathrm{MAP}} = \argmax_{\mathbf{X}}\prod_{i=1}^{n}f_i(\mathbf{X}_i).
\end{equation}
Each factor is a function of the form
\begin{equation}
  \label{eq:factor_function}
  f_i(\mathbf{X}_i) = \mathrm{exp}\left(-\frac{1}{2}\lVert \mathbf{e}_i \rVert_{\bm{\Sigma}_i}^2\right),
\end{equation}
where $\mathbf{e}_i = h(\mathbf{X}_i)-\mathbf{z}_i$ is the error vector with $h(\mathbf{X}_i)$ being the prediction function, and $\mathbf{z}_i$ the actual measured value. 
The Mahalanobis norm of an error vector $\mathbf{e}$ with the covariance matrix $\mathbf{\Sigma}$ is defined as $\lVert\mathbf{e}\rVert_{\bm{\Sigma}}^2=\mathbf{e}^{\top}\mathbf{\Sigma}^{-1}\mathbf{e}$.
Substituting \autoref{eq:factor_function} into \autoref{eq:map_max} yields:
\begin{equation}
  \label{eq:map_max_exp}
  \mathbf{X}^{\mathrm{MAP}} = \argmax_{\mathbf{X}} \mathrm{exp}\left(-\frac{1}{2}\sum_{i=1}^{n}\lVert \mathbf{e}_i \rVert_{\bm{\Sigma}_i}^2\right).
\end{equation}
By taking the negative log-likelihood, we can transform the \ac{map} maximization problem in \autoref{eq:map_max_exp} into an equivalent nonlinear least-squares minimization:
\begin{equation}
  \label{eq:min_nonlinear}
  \mathbf{X}^{\mathrm{MAP}} = \argmin_{\mathbf{X}}\sum_{i=1}^{n}\lVert \mathbf{e}_i \rVert_{\bm{\Sigma}_i}^2.
\end{equation}

Non-linear optimization methods, such as \ac{lm}~\cite{more2006levenberg}, solve the non-linear objective function by iteratively solving its linearized approximation:
\begin{equation}
  h(\mathbf{X}_i) \approx h(\mathbf{X}_i^0) + \mathbf{H}_i \bm{\Delta}_i,
\end{equation}
where $\bm{\Delta}_i = \mathbf{X}_i - \mathbf{X}_i^0$ is the state update vector and $\mathbf{H}_i$ is the Jacobian of $h$ at the linearization point $\mathbf{X}_i^0$:
\begin{equation}
  \mathbf{H}_i = \left. \frac{\partial h(\mathbf{X}_i)}{\partial \mathbf{X}_i}\right|_{\mathbf{X}_i^0}.
\end{equation}

\subsection{Optimization on manifold}
\label{section:manifold_optimization}

The optimization problem \autoref{eq:min_nonlinear} on the $\mathcal{S}$ manifold is solved by iteratively optimizing a reparametrized problem in the tangent space around the current solution guess $\bar{\mathbf{x}}$
\begin{equation}
  \bm{\tvec}^{*} = \argmin_{\mathbf{\bm{\xi}}\in \mathbb{R}^{12}}\sum_{i=1}^n\lVert \mathrm{e}_i(\mathcal{R}_{\bar{\mathbf{x}}}(\bm{\tvec})) \rVert^2,
\end{equation}
 followed by a solution guess update 
\begin{equation}
  \bar{\mathbf{x}} \leftarrow \mathcal{R}_{\bar{\mathbf{x}}}(\bm{\tvec}^{*}).
\end{equation}


\section{UAV Dynamics Model}
\label{section:mas_model}
The \ac{uav} is controlled by changing the rotational speed of each propeller, which produces thrust.
We first show how we obtain linear acceleration and angular acceleration from forces in torque, and then we introduce the \ac{uav} motion equations.

\subsection{Forces and Torques}
\label{section:uav_dynamics}

The translational dynamics of a multirotor \ac{uav} are characterized by the forces $\mathbf{f}^{\mathcal{W}}$ in the world frame $\mathcal{W}$:
\begin{equation}
  \label{eq:trans_dynamics}
      \mathbf{f}^{\mathcal{W}} =
      \mathbf{R}^\mathcal{W}_\mathcal{B}\mathbf{f}^{\mathcal{B}}_t + 
      \mathbf{f}^\mathcal{W}_g,
\end{equation}
where $\mathbf{R}_{\mathcal{B}}^{\mathcal{W}}$ is the rotation of the \ac{uav} body frame in the world frame, $\mathbf{f}^{\mathcal{B}}_t$ is the \textit{thrust force} produced by the propellers of the \ac{uav},
$\mathbf{f}^\mathcal{W}_g = -mg\mathbf{e}^\mathcal{W}_3$ is the force produced by Earth's gravitational acceleration (with  magnitude $g=\SI{9.81}{\meter\per\second\squared}$) acting along the $\mathbf{e}^{\mathcal{W}}_3=\left[\begin{matrix}0 & 0 & 1\end{matrix}\right]^T$ axis of the world frame.

The rotational dynamics are expressed by the torques $\bm{\tau}^{\mathcal{B}}$ in the body frame $\mathcal{B}$:
\begin{equation}
  \label{eq:rot_dynamics}
  \bm{\tau}^{\mathcal{B}} = 
  \mathbf{M}\bm{\alpha}^\mathcal{B} + 
  \bm{\omega}^\mathcal{B}\times\mathbf{M}\bm{\omega}^\mathcal{B},
\end{equation}
where $\mathbf{M}\in\mathbb{R}^3$ is the inertia matrix,
$\bm{\alpha}^{\mathcal{B}}$ and $\bm{\omega}^{\mathcal{B}}$ are the body frame angular acceleration and angular velocity, respectively.

The forces and torques that act upon the \ac{uav} are caused by the aerodynamic effects of spinning propellers. 
The \textit{thrust force} and \textit{drag torque} produced by the propeller $i$ are:
\begin{equation}
  \mathbf{f}_i^{\mathcal{P}_i} = c_f\massquared\mathbf{e}^{\mathcal{P}_i}_3,
\end{equation}
\begin{equation}
  \bm{\tau}_{d_i}^{\mathcal{P}_i} = \left(-1\right)^{\left(i\right)}c_d\massquared\mathbf{e}^{\mathcal{P}_i}_3,
\end{equation}
where $c_f$ and $c_d$ are the thrust force and drag torque coefficients, which map the signed square of the rotor angular velocity to a force and torque, respectively. 

To obtain the total $\textit{thrust force}$ and $\textit{drag torque}$ in a body frame, we sum them up as:
\begin{equation}
  \mathbf{f}^{\mathcal{B}}_t = \sum_{i=1}^n\mathbf{R}_{\mathcal{P}_i}^\mathcal{B}\mathbf{f}_i^{\mathcal{P}_i},\quad
  \bm{\tau}_d^{\mathcal{B}} = \sum_{i=1}^n\mathbf{R}_{\mathcal{P}_i}^\mathcal{B}\bm{\tau}_{d_i}^{\mathcal{P}_i}.
\end{equation}
Since $\mathbf{f}_i^{\mathcal{P}_i}$ are not applied in the body frame, they cause the \textit{thrust torque}:
\begin{equation}
  \bm{\tau}_{f}^{\mathcal{B}} =
  \sum_{i=1}^n\bm{\rho}_i^{\mathcal{B}}\times\mathbf{R}_{\mathcal{P}_i}^\mathcal{B}\mathbf{f}_i^{\mathcal{P}_i},
\end{equation}
where $\bm{\rho}_i^\mathcal{B}$ is the position of rotor $i$ in the body frame.
The \textit{total torque} obtained is the sum
\begin{equation}
  \bm{\tau}^{\mathcal{B}} =
\bm{\tau}_{d}^{\mathcal{B}} +
\bm{\tau}_{f}^{\mathcal{B}}.
\end{equation}
For underactuated \acp{uav}, we assume $\mathbf{R}_{\mathcal{P}_i}^\mathcal{B}\approx \mathbf{I}$ (i.e. all propellers lie on the same plane), therefore $\bm{\tau}_{d}^{\mathcal{B}}$ acts mostly along the $\mathbf{e}_3^{\mathcal{B}}$ axis (causing rotation in yaw) and $\bm{\tau}_{f}^{\mathcal{B}}$ acts along the $\mathbf{e}_1^{\mathcal{B}}$ and $\mathbf{e}_2^{\mathcal{B}}$ axes (causing rotation in roll and pitch).

We need the body-frame linear acceleration $\mathbf{a}^\mathcal{B}$ and the angular acceleration $\bm{\alpha}^\mathcal{B}$ for the propagation of the \ac{uav} state according to its equations of motion.
Using equations \autoref{eq:trans_dynamics} and \autoref{eq:rot_dynamics}, we can obtain $\mathbf{a}^\mathcal{B}$ and $\bm{\alpha}^\mathcal{B}$ from known $\mathbf{f}^{\mathcal{B}}_t$ and $\bm{\tau}^{\mathcal{B}}$:
\begin{equation}
  \label{eq:lin_acc}
  \mathbf{a}^\mathcal{B} = 
      \frac{\mathbf{f}^{\mathcal{B}}_t + \mathbf{R}^\mathcal{B}_\mathcal{W} \mathbf{f}^\mathcal{W}_g}{m},
\end{equation}
\begin{equation}
  \label{eq:ang_acc}
  \bm{\alpha}^\mathcal{B} =
  \mathbf{M}^{-1}\left(\bm{\tau}^{\mathcal{B}} - \bm{\omega}^\mathcal{B}\times\mathbf{M}\bm{\omega}^\mathcal{B}\right).
\end{equation}

Apart from \textit{thrust force}, \textit{drag torque}, and \textit{thrust torque}, the motion of the \ac{uav} is also influenced by other forces and moments \cite{bouabdallah2007full}, such as \textit{aerodynamic drag force}, \textit{hub force}, \textit{rolling moment}, \textit{ground effect}, and \textit{blade flapping moment}~\cite{svacha2020imu}.
We do not consider these effects in the model, as they are orders of magnitude smaller than the modeled components.

\subsection{Equations of Motion}
\label{section:motion_model}

To describe the evolution of \ac{uav} state $\mathbf{x} \in \mathcal{S}$ in time, we will define the \textit{tangent vector} at $\mathbf{x}_{t_0}$ as the derivative of a trajectory $\mathbf{x}(t)$ at $\mathbf{x}_{t_0}=\mathbf{x}(t_0)$:
\begin{multline}
  \dot{\mathbf{x}}(t) \triangleq \\ 
  \left[\dot{\mathbf{p}}^{\mathcal{W}}(t,\mathbf{x}_{t_0}), \dot{\mathbf{R}}^{\mathcal{W}}_{\mathcal{B}}(t,\mathbf{x}_{t_0}), \dot{\mathbf{v}}^{\mathcal{W}}(t,\mathbf{x}_{t_0}), \dot{\bm{\omega}}^{\mathcal{W}}(t,\mathbf{x}_{t_0})\right] \in \mathbb{R}^{12},
\end{multline}
where
\begin{equation}
  \label{eq:motion_eq}
\begin{aligned}
  \dot{\mathbf{p}}^\mathcal{W}(t,\mathbf{x}_{t_0}) &= \mathbf{v}^\mathcal{W}(t), \\
  \dot{\mathbf{R}}^\mathcal{W}_\mathcal{B}(t,\mathbf{x}_{t_0}) &= \left[\bm{\omega}^\mathcal{W}(t)\right]_{\times}, \\
  \dot{\mathbf{v}}^\mathcal{W}(t,\mathbf{x}_{t_0}) &= \mathbf{R}^\mathcal{W}_\mathcal{B}(t) \mathbf{a}^{\mathcal{B}}(t), \\
  \dot{\bm{\omega}}^\mathcal{W}(t,\mathbf{x}_{t_0}) &= \mathbf{R}^\mathcal{W}_\mathcal{B}(t) \bm{\alpha}^\mathcal{B}(t).
\end{aligned}
\end{equation}


\subsection{Noise model}
\label{section:noise_model}

We model the errors in linear and angular accelerations as 
\begin{equation}
\begin{aligned}
  \label{eq:noise_model}
  \mathbf{\hat{a}}^{\mathcal{B}} &= \mathbf{a}^{\mathcal{B}} + \bm{\eta}_{\mathbf{a}}^{\mathcal{B}} - \bm{\nu}_{\mathbf{a}}^{\mathcal{B}} - \bm{\kappa}_{\mathbf{a}}^{\mathcal{B}}, \\
  \bm{\hat{\alpha}}^{\mathcal{B}} &= \bm{\alpha}^{\mathcal{B}} + \bm{\eta}_{\bm{\alpha}}^{\mathcal{B}} - \bm{\nu}_{\bm{\alpha}}^{\mathcal{B}} - \bm{\kappa}_{\bm{\alpha}}^{\mathcal{B}},
\end{aligned}
\end{equation}
where $\mathbf{\hat{a}}^{\mathcal{B}}$ and $\bm{\hat{\alpha}}^{\mathcal{B}}$ are linear and angular accelerations obtained from \autoref{eq:lin_acc} and \autoref{eq:ang_acc}, $\bm{\eta}$ is the zero-mean Gaussian noise modeling \ac{mas} measurement noise. 
The random walk bias $\bm{\nu}$ encompasses nonlinearities of \ac{mas} to thrust/torque mapping, and unmodeled forces/torques, as well as external disturbances, such as wind.
The last term, $\bm{\kappa}$, is the constant offset, which accounts for non-coinciding rotation planes of individual propellers, inaccuracies in \ac{uav} mass and inertia matrix, and misalignment between the center of gravity and the geometric center of the \ac{uav}.



\section{MAS Preintegration}
\label{section:mas_preintegration}

The full state of the \ac{uav} can be obtained by propagating $\mathbf{\hat{a}}^{\mathcal{B}}$ and $\bm{\hat{\alpha}}^{\mathcal{B}}$ through the equations of motion~\autoref{eq:motion_eq}.
However, integrating the accelerations every time the \ac{mas} is measured would be very inefficient, as a factor would be created for every measurement and the optimized graph would be too large.
Moreover, after receiving a delayed measurement from a past time, all \ac{mas} measurements would have to be integrated again up to the current time.
To avoid repeated integration of the same measurements, we compute only the relative motion of the preintegrated \ac{mas} measurements in the local coordinates, which does not depend on the initial state.

\subsection{State Prediction}
Let us assume a state of the \ac{uav} in a time instant $i$:
\begin{equation}
  \mathbf{x}_i = \left\{\mathbf{p}_i, \mathbf{R}_i, \mathbf{v}_i, \bm{\omega}_i\right\} \in \mathcal{S}.
\end{equation}

Let us consider another \ac{uav} state $\hat{\mathbf{x}}_j$, which we want to predict based on the previous state $\mathbf{x}_i$ and the local coordinate vector $\bm{\tvec}(t_{ij})$ integrated from zero to time $\tij = t_j - t_i$ in the tangent space:
\begin{equation}
  \label{eq:preintegrated_vector}
  \bm{\tvec}(\tij) = [\mathbf{p}(\tij), \bm{\theta}(\tij), \mathbf{v}(\tij), \bm{\omega}(\tij)]^{\top} \in \mathbb{R}^{12},
\end{equation}
where
\begin{equation}
\begin{aligned}
  \mathbf{p}(\tij) &= \mathbf{R}_i^{\top}\mathbf{v}_i t_{ij} + 0.5\mathbf{R}_i^{\top}\mathbf{g}t^2_{ij} +  \mathbf{p}_v(\tij), \\
  \bm{\theta}(\tij) &= \mathbf{R}^{\top}_i\bm{\omega}_i\tij + \bm{\theta}_{\omega}(\tij), \\
  \mathbf{v}(\tij) &= \mathbf{R}^{\top}_i\mathbf{g}\tij + \mathbf{v}_a(\tij), \\
  \bm{\omega}(\tij) &= \bm{\omega}_{\alpha}(\tij).
\end{aligned}
\end{equation}
To obtain $\mathbf{p}_v(\tij)$, $\bm{\theta}_{\omega}(\tij)$, $\mathbf{v}_a(\tij)$, and $\bm{\omega}_{\alpha}(\tij)$, we will need to solve the differential equations
\begin{equation}
  \label{eq:diff_eq}
\begin{aligned}
  \dot{\mathbf{p}}_v(\tij) &= \mathbf{v}_a(\tij), \\
  \dot{\bm{\theta}}_\omega(\tij) &= \bm{\omega}_{\alpha}(\tij), \\
  \dot{\mathbf{v}}_a(\tij) &= \mathbf{R}_{\mathcal{B}}^i(\tij)\mathbf{a}^{\mathcal{B}}(\tij), \\
  \dot{\bm{\omega}}_\alpha(\tij) &= \mathbf{J}(\bm{\theta}(\tij))^{-1}\bm{\alpha}^{\mathcal{B}}(\tij),
\end{aligned}
\end{equation}
where $\mathbf{R}_{\mathcal{B}}^i(\tij) = \mathrm{exp}(\left[\bm{\theta}(\tij)\right]_{\times})$ is the time-evolving orientation of the \ac{uav} body frame with respect to $\mathbf{x}_i$.
The Jacobian inverse $\mathbf{J}(\bm{\theta}(\tij))^{-1}$ models the incremental changes of $\bm{\alpha}^{\mathcal{B}}(\tij)$ in the local coordinates and $\tij$ indicates preintegration time from $t_i$ to $t_j$.
We will get back to solving~\autoref{eq:diff_eq} in~\autoref{section:discrete_state_vector}. 
For now, we assume that we have the solution and thus we can predict the state $\hat{\mathbf{x}}_j$ by retracting the local coordinate vector $\bm{\xi}(t_{ij})$ from~\autoref{eq:preintegrated_vector} back to the \ac{uav} state manifold:
\begin{equation}
  \label{eq:predict_state}
  \hat{\mathbf{x}}_j = \mathcal{R}_{\mathbf{x}_i}(\bm{\tvec}(\tij)) = \{\hat{\mathbf{p}}_j, \hat{\mathbf{R}}_j, \hat{\mathbf{v}}_j, \hat{\bm{\omega}}_j\} \in {\mathcal{S}},
\end{equation}
where
\begin{equation}
\begin{aligned}
  \hat{\mathbf{p}}_j &= \mathbf{p}_i + \mathbf{R}_i(\mathbf{R}^{\top}_i \mathbf{v}_i\tij + 0.5\mathbf{R}^{\top}_i \mathbf{g}t^2_{ij} + \mathbf{p}_v(\tij)), \\
  \hat{\mathbf{R}}_j &= \mathbf{R}_i\mathrm{exp}([\mathbf{R}^{\top}_i \bm{\omega}_i\tij + \bm{\theta}_{\omega}(\tij)]_{\times}), \\
  \hat{\mathbf{v}}_j &= \mathbf{v}_i + \mathbf{R}_i (\mathbf{R}^{\top}_i\mathbf{g}\tij + \mathbf{v}_a(\tij)), \\
  \hat{\bm{\omega}}_j &= \bm{\omega}_i + \mathbf{R}_i\bm{\omega}_{\alpha}(\tij). \\
\end{aligned}
\end{equation}

\subsection{Preintegrated Discrete State Vector}
\label{section:discrete_state_vector}
As the accelerations are measured in discrete steps, we find the preintegrated discrete state vector
\begin{equation}
  \bm{\chi}_{[k+1]} = [\mathbf{p}_{v[k+1]}, \bm{\theta}_{\omega[k+1]}, \mathbf{v}_{a[k+1]}, \bm{\omega}_{\alpha[k+1]}]^{\top},
\end{equation}
as a discrete solution to the differential equations \autoref{eq:diff_eq} using the Euler method $\mathbf{x}_{[k+1]} = \mathbf{x}_{[k]} + f(t_{[k]}, \mathbf{x}_{[k]})\Delta t$:
\begin{equation}
\begin{aligned}
  \label{eq:int_discrete}
  \mathbf{p}_{v[k+1]} &= \mathbf{p}_{v[k]} +\mathbf{v}_{a[k]}\Delta t + 0.5\mathbf{R}_{[k]}\hat{\mathbf{a}}_{[k]}^{\mathcal{B}}\Delta t^2, \\
  \bm{\theta}_{\omega[k+1]} &= \bm{\theta}_{\omega[k]} +\bm{\omega}_{\alpha[k]}\Delta t + 0.5\mathbf{J}(\bm{\theta}_{\omega[k]})^{-1}\hat{\bm{\alpha}}_{[k]}^{\mathcal{B}}\Delta t^2, \\
  \mathbf{v}_{a[k+1]} &= \mathbf{v}_{a[k]} + \mathbf{R}_{[k]} \hat{\mathbf{a}}_{[k]}^{\mathcal{B}} \Delta t, \\
  \bm{\omega}_{\alpha[k+1]} &= \bm{\omega}_{\alpha[k]} + \mathbf{J}(\bm{\theta}_{\omega[k]})^{-1} \hat{\bm{\alpha}}_{[k]}^{\mathcal{B}} \Delta t,
\end{aligned}
\end{equation}
where $\mathbf{R}_{[k]}=\mathrm{exp}\left(\left[\bm{\theta}_{\omega[k]}\right]_{\times}\right)$. 
The measurements $\hat{\mathbf{a}}_{[k]}^{\mathcal{B}}$ and $\hat{\bm{\alpha}}_{[k]}^{\mathcal{B}}$ include the noise, bias, and initial offset from \autoref{eq:noise_model}.

From \cref{eq:int_discrete}, we can see that the integrated quantities are a function of the state from the previous preintegration step
\begin{equation}
  \label{eq:chi_nonlinear_f}
  \bm{\chi}_{[k+1]} = f \left(\mathbf{p}_{v[k]}, \bm{\theta}_{\omega[k]}, \mathbf{v}_{a[k]}, \bm{\omega}_{\alpha[k]}\right),
\end{equation}
and are not dependent on the initial state $\mathbf{x}_i$, which is an important property that allows for the existence of an independent \ac{mas} factor.

\section{MAS Factors}
\label{section:mas_factors}

\subsection{MAS Binary Factor}
\label{section:mas_binary_factor}
The proposed MAS factor is a binary factor, i.e., it constrains two consecutive \ac{uav} states $\mathbf{x}_i$, and $\mathbf{x}_j$ using the measurements $\mathbf{a}_i$, $\bm{\alpha}_i$ together with their estimated biases $\mathbf{b}_i^a$ and $\mathbf{b}_i^{\alpha}$, respectively. Formally, this is written as:
\begin{equation}
  f_{ij}^{\mathrm{MAS}}(\mathbf{x}_i, \mathbf{x}_j, \mathbf{a}_i, \bm{\alpha}_i, \mathbf{b}_i^a, \mathbf{b}_i^{\alpha}).
\end{equation}

The residual vector that we want to minimize is computed as the difference between the actual state and the predicted state:
\begin{equation}
  \mathbf{e}_i = \mathbf{x}_j \ominus \hat{\mathbf{x}}_j,
\end{equation}
where $\ominus$ denotes the manifold-minus operator and $\hat{\mathbf{x}}_j$ is the state predicted by the function
\begin{equation}
  \hat{\mathbf{x}}_j = h(\mathbf{x}_i, \bm{\tvec}_{ij}),
\end{equation}
which is defined in~\autoref{eq:predict_state}.

Further, we need the Jacobian of the prediction function for the gradient descent of \ac{lm}:
\begin{multline}
  \mathbf{H}_i = \left. \frac{\partial h(\mathbf{x}_i, \bm{\tvec}_{ij})}{\partial \mathbf{x}_i}\right|_{\mathbf{x}_i^0}
  =
  \left[
  \begin{smallmatrix}
    \mathbf{I}_3 && \mathbf{p}_v(t_{ij}) && \mathbf{0}_3 && \mathbf{0}_3 \\
    \mathbf{0}_3 && \mathrm{exp}(\left[\bm{\omega}_i t_{ij}\right]_{\times} && \mathbf{0}_3 && \mathbf{0}_3 \\
    \mathbf{0}_3 && \mathbf{v}_a(t_{ij}) && \mathbf{I}_3 && \mathbf{0}_3 \\
    \mathbf{0}_3 && \bm{\omega}_{\alpha}(t_{ij}) && \mathbf{0}_3 && \mathbf{I}_3
  \end{smallmatrix}
  \right],
\end{multline}

\subsection{MAS Factor Noise Propagation}
\label{section:mas_noise_propagation}

To minimize \autoref{eq:min_nonlinear}, we need to obtain the covariance $\bm{\Sigma}_i$, which is updated with each preintegration step:
\begin{equation}
  \bm{\Sigma}_{[k+1]} = \mathbf{A}_{[k]}\bm{\Sigma}_{[k]}\mathbf{A}_{[k]}^{\top} + \mathbf{B}_{[k]}\bm{\Sigma}_{\eta}^{a}\mathbf{B}_{[k]}^{\top} + \mathbf{C}_{[k]}\bm{\Sigma}_{\eta}^{\alpha}\mathbf{C}_{[k]}^{\top},
\end{equation}
where $\bm{\Sigma}_{\eta}^{a}$ and $\bm{\Sigma}_{\eta}^{\alpha}$ are the covariance matrices of the linear and angular acceleration, respectively.
$\mathbf{A}_{[k]}$ is the matrix of partial derivatives of the nonlinear function $f$ from \autoref{eq:chi_nonlinear_f} w.r.t. $\bm{\chi}_{[k]}$, matrix $\mathbf{B}_{[k]}$ contains the partial derivatives w.r.t. $\mathbf{a}_{[k]}^{\mathcal{B}}$, and partial derivatives w.r.t. $\bm{\alpha}_{[k]}^{\mathcal{B}}$ reside in $\mathbf{C}_{[k]}$:
\begin{gather}
  \mathbf{A}_{[k]}=  
  \left[
\begin{smallmatrix}
  \frac{\partial\mathbf{p}_{[k+1]}}{\partial\mathbf{p}_{[k]}} && \frac{\partial\mathbf{p}_{[k+1]}}{\partial\bm{\theta}_{[k]}} && \frac{\partial\mathbf{p}_{[k+1]}}{\partial\mathbf{v}_{[k]}} && \frac{\partial\mathbf{p}_{[k+1]}}{\partial\bm{\omega}_{[k]}} \\
  \frac{\partial\bm{\theta}_{[k+1]}}{\partial\mathbf{p}_{[k]}} && \frac{\partial\bm{\theta}_{[k+1]}}{\partial\bm{\theta}_{[k]}} && \frac{\partial\bm{\theta}_{[k+1]}}{\partial\mathbf{v}_{[k]}} && \frac{\partial\bm{\theta}_{[k+1]}}{\partial\bm{\omega}_{[k]}} \\
  \frac{\partial\mathbf{v}_{[k+1]}}{\partial\mathbf{p}_{[k]}} && \frac{\partial\bm{\theta}_{[k+1]}}{\partial\mathbf{v}_{[k]}} && \frac{\partial\mathbf{v}_{[k+1]}}{\partial\mathbf{v}_{[k]}} && \frac{\partial\mathbf{v}_{[k+1]}}{\partial\bm{\omega}_{[k]}} \\
  \frac{\partial\bm{\omega}_{[k+1]}}{\partial\mathbf{p}_{[k]}} && \frac{\partial\bm{\omega}_{[k+1]}}{\partial\bm{\theta}_{[k]}} && \frac{\partial\bm{\omega}_{[k+1]}}{\partial\mathbf{v}_{[k]}} && \frac{\partial\bm{\omega}_{[k+1]}}{\partial\bm{\omega}_{[k]}} 
\end{smallmatrix}
  \right] \nonumber
\\
=
  \left[
\begin{smallmatrix}
  \mathbf{I}_3 && \frac{1}{2}\mathbf{R}_{[k]}[\mathbf{a}^{\mathcal{B}}_{[k]}]_{\times}\mathbf{J}(\bm{\theta}_{\omega[k]})\Delta t^2 && \mathbf{I}_3\Delta t && \mathbf{0}_3 \\
  \mathbf{0}_3 && \mathbf{I}_3 - \frac{1}{4} [\bm{\alpha}^{\mathcal{B}}_{[k]}]_{\times} \Delta t^2 && \mathbf{0}_3 && \mathbf{I}_3 \Delta t \\
  \mathbf{0}_3 && \mathbf{R}_{[k]}[\mathbf{a}^{\mathcal{B}}_{[k]}]_{\times}\mathbf{J}(\bm{\theta}_{\omega[k]}) \Delta t && \mathbf{I}_3 && \mathbf{0}_3 \\
  \mathbf{0}_3 && -\frac{1}{2} [\bm{\alpha}^{\mathcal{B}}_{[k]}]_{\times} \Delta t && \mathbf{0}_3 && \mathbf{I}_3
\end{smallmatrix}
  \right],
\end{gather}

\begin{equation}
  \mathbf{B}_{[k]}=  
  \left[
\begin{smallmatrix}
  \frac{\partial\mathbf{p}_{[k+1]}}{\partial\mathbf{a}_{[k]}} \\
  \frac{\partial\bm{\theta}_{[k+1]}}{\partial\mathbf{a}_{[k]}} \\
  \frac{\partial\mathbf{v}_{[k+1]}}{\partial\mathbf{a}_{[k]}} \\
  \frac{\partial\bm{\omega}_{[k+1]}}{\partial\mathbf{a}_{[k]}} 
\end{smallmatrix}
  \right]
=
  \left[
\begin{smallmatrix}
  \frac{1}{2}\mathbf{R}_{[k]} \Delta t^2 \\
  \mathbf{0}_3 \\
  \mathbf{R}_{[k]} \Delta t \\
  \mathbf{0}_3 
\end{smallmatrix}
  \right],
\end{equation}

\begin{equation}
  \mathbf{C}_{[k]}=  
  \left[
\begin{smallmatrix}
  \frac{\partial\mathbf{p}_{[k+1]}}{\partial\bm{\alpha}_{[k]}} \\
  \frac{\partial\bm{\theta}_{[k+1]}}{\partial\bm{\alpha}_{[k]}} \\
  \frac{\partial\mathbf{v}_{[k+1]}}{\partial\bm{\alpha}_{[k]}} \\
  \frac{\partial\bm{\omega}_{[k+1]}}{\partial\bm{\alpha}_{[k]}} 
\end{smallmatrix}
  \right]
=
  \left[
\begin{smallmatrix}
  \mathbf{0}_3 \\
  \frac{1}{2}\mathbf{J}(\bm{\theta}_{[k]})^{-1} \Delta t^2 \\
  \mathbf{0}_3 \\
  \mathbf{J}(\bm{\theta}_{[k]})^{-1} \Delta t \\
\end{smallmatrix}
  \right].
\end{equation}

\subsection{MAS Bias Binary Factor}
\label{section:mas_bias_binary_factor}
  The random walk bias introduced in \autoref{section:noise_model} forms a binary factor
  \begin{equation}
    f^{\mathrm{BIAS}}(\mathbf{b}_i, \mathbf{b}_j, \bm{\eta}^{ba}, \bm{\eta}^{b\alpha}),
  \end{equation} 
  which models the evolution of linear and angular acceleration bias $\mathbf{b} = [\mathbf{b}^a, \mathbf{b}^{\alpha}]^{\top}$ in time, with their initial values modeling the constant offsets.
  To model the random walk, we integrate a zero-mean Gaussian noise
  \begin{equation}
    \mathbf{b}^a_j = \mathbf{b}^a_i + \bm{\eta}^{ba}, \qquad \bm{\eta}^{ba} = \mathcal{N}(\mathbf{0}; \bm{\Sigma}^{ba}),
  \end{equation} 
  and
  \begin{equation}
    \mathbf{b}^{\alpha}_j = \mathbf{b}^{\alpha}_i + \bm{\eta}^{b\alpha}, \qquad \bm{\eta}^{b\alpha} = \mathcal{N}(\mathbf{0}; \bm{\Sigma}^{b\alpha}).
  \end{equation}
  The least squares error terms to be optimized are
  \begin{equation}
    \lVert\mathbf{e}_i\rVert^2 = \lVert \mathbf{b}^a_j - \mathbf{b}^a_i\rVert_{\bm{\Sigma}^{ba}}^2 + \lVert \mathbf{b}^{\alpha}_j - \mathbf{b}^\alpha_i\rVert_{\bm{\Sigma}^{b\alpha}}^2,
  \end{equation}
  with corresponding Jacobian matrices
  \begin{equation}
    \mathbf{H}^{ba}_i = \left[ \begin{smallmatrix} - \mathbf{I}_3 \\ \mathbf{0}_3\end{smallmatrix} \right], \qquad
    \mathbf{H}^{b\alpha}_i = \left[ \begin{smallmatrix} \mathbf{0}_3 \\ - \mathbf{I}_3\end{smallmatrix} \right].
  \end{equation}


\section{MAS-LO}
\label{section:maslo}

\begin{figure}[thpb]
  \centering
  \adjincludegraphics[width=1.0\linewidth, trim={{0.05\width} {0.10\height} {0.06\width} {0.07\height}}, clip]{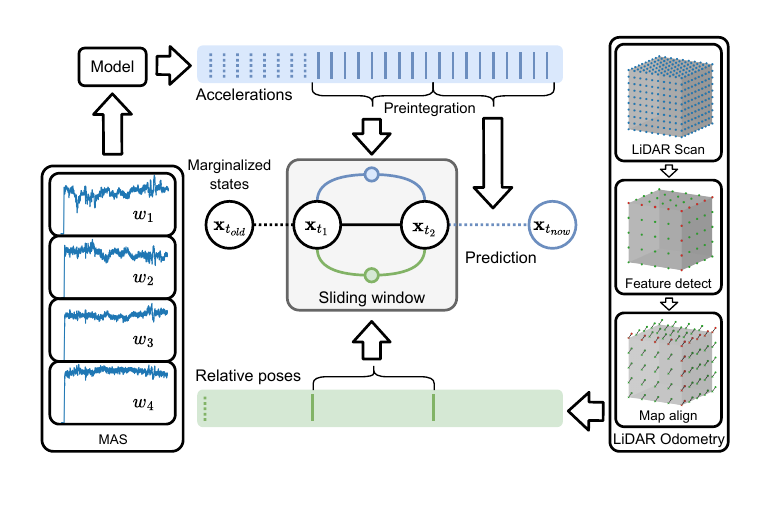}
  \caption{
    The \ac{mas} measurements are passed through the propulsion model to obtain accelerations that serve two purposes after preintegration; if a \ac{lo} measurement becomes available, \ac{mas} factor is created. Otherwise, the preintegrated accelerations are used to propagate the state to the current time.
    The \ac{lo} method extracts edge and plane features from a \ac{lidar} scan and matches them into a local feature map to obtain relative transformations that generate a \ac{lo} factor.
    The states that are older than the fixed-length sliding window are marginalized out.
    }
  \label{fig:factor_graph}
\end{figure}

  In this section, we present \ac{maslo}, an \ac{mas}-based localization algorithm.
  The \ac{mas} measurements cannot be used on their own to estimate the \ac{uav} state due to the large drift that results from the double integration of accelerations.
  To constrain the drift we combine the preintegrated \ac{mas} factor with \ac{lidar} odometry factor (see~\autoref{fig:factor_graph}), which makes the acceleration bias observable.
  The \ac{lidar} odometry is adopted from the \ac{liosam} \cite{shan2020lio} algorithm, which will facilitate the evaluation by focusing on the comparison of \ac{mas} and \ac{imu} preintegration methods, while keeping the same \ac{lidar} odometry method.

\subsection{Overview}
\label{section:maslo_overview}

Our approach is based on the factor graph formulation with three kinds of factors: \ac{mas} binary factor (\autoref{section:mas_binary_factor}), \ac{mas} bias binary factor (\autoref{section:mas_bias_binary_factor}), and \ac{lo} binary factor (\autoref{section:lo_binary_factor}).
The \ac{mas} measurements are continuously preintegrated. When the \ac{lidar} outputs a new scan, the scan is matched into a sliding window feature map to obtain a transformation that forms the \ac{lo} factor.
At the same time, the preintegrated \ac{mas} measurements form a \ac{mas} factor.
  The \ac{isam2}\cite{kaess2012isam2} solver minimizes the residual errors of the factors inside the sliding window, and marginalizes the states outside this window.
  Finally, the \ac{mas} measurements arriving after the \ac{lidar} scan are preintegrated up to the current time to predict the current state.

\subsection{Feature Extraction}
\label{section:feature_extraction}

\ac{lidar} scans are typically too large for real-time scan matching of all points.
Moreover, a large part of the points are redundant and do not contribute to the correct alignment of two scans.
In fact, an uneven distribution of redundant points throughout the environment can even prevent convergence in the case of a geometrically degenerate environment~\cite{petracek2024rms}.

To adhere to the real-time constraints and reduce drift, spatially distributed geometrical features are detected for correspondence matching instead of raw points matching.
First, the approach of~\cite{zhang2017low} is used to estimate the smoothness of each point. Then, a fixed number of points with the lowest and highest smoothness are selected as the edge and plane feature points, respectively.

\subsection{Scan Matching}
\label{section:scan_matching}

A sliding window of features detected in previous scans is kept in a voxelized map in the world frame to match the current scan. 
After detecting features in the current scan, they are transformed to the world frame. Correspondences are found by searching for the closest edge line calculated from the edge points in the sliding window. Similarly, the closest planar patch is found for each planar feature point.
The distances of point-to-line and point-to-plane matches then form the residuals in nonlinear optimization problem solved by the \ac{lm} method.
When the \ac{lm} converges, the obtained $SE(3)$ transform forms the \ac{lo} factor.


\subsection{LO Binary Factor}
\label{section:lo_binary_factor}

  The scan matching provides the relative transformation between two poses, i.e. the \ac{lo} factor constrains the two poses (not full states) $\mathbf{s}_i, \mathbf{s}_j \in SE(3)$:
  \begin{equation}
    f^{LO}(\mathbf{s}_i, \mathbf{s}_j, \bm{\Delta}\mathbf{x}_{ij}, \bm{\eta}^p, \bm{\eta}^{\theta}),
  \end{equation}
  with the prediction function
  \begin{equation}
    h(\mathbf{x}_i) = \left[\mathbf{R}_i^{\top}(\mathbf{p}_j - \mathbf{p}_i), \mathbf{R}_j \ominus \mathbf{R}_i\right]^{\top},
  \end{equation}
  its Jacobian:
\begin{equation}
  \mathbf{H}_i = 
  \left[
  \begin{smallmatrix}
    -\mathbf{I}_3 && [\mathbf{p}_i - \mathbf{p}_j]_{\times} \\
    \mathbf{0}_3 && \mathbf{J}^{-1}(\bm{\theta}_{ij}) \\
  \end{smallmatrix}
  \right],
\end{equation}
  and measurement
  \begin{equation}
    \mathbf{z}_i = [\bm{\Delta}\mathbf{p}_{ij}, \bm{\Delta}\bm{\theta}_{ij}]^{\top}.
  \end{equation}
  

\subsection{GTSAM implementation}
\label{section:gtsam}

  We have implemented the \ac{mas} preintegration, the \ac{mas} factor, and the \ac{maslo} localization algorithm using the open-source GTSAM framework \cite{dellaert2022gtsam}.
  GTSAM is a C++ library developed for modeling and solving complex estimation problems, such as \ac{slam} and \ac{sfm}, by searching for a maximum a posteriori probability.
  The algorithms implemented in GTSAM exploit the sparsity of the optimization problems to be computationally efficient.
  The framework also facilitates creating custom factors, their residual functions, and Jacobians by providing tools to work with manifolds, tangent spaces, and Lie groups.
  Large factor graphs require an efficient solver that exploits the sparse structure of the problem to run in real-time.
  The \ac{isam2}\cite{kaess2012isam2} solver implemented in GTSAM allows for fast execution times by operating on the Bayes tree data structure \cite{kaess2011bayes} with iterative local relinearization of variables affected by a new measurement.
  
  Developing our algorithms inside the GTSAM framework also allows that any inertial localization algorithm for \acp{uav} using GTSAM can be modified to use the \ac{mas} factor instead of the default \ac{imu} factor.


\section{Evaluation and experiments}
\label{section:evaluation}

The \ac{mas} preintegration and \ac{mas} factor as a part of \ac{maslo} localization algorithm are evaluated on six recorded datasets, which we also provide with open access.
We evaluated the proposed solution in terms of noise, runtime, and, most importantly, trajectory error w.r.t. \ac{rtk} ground truth.

\subsection{Hardware setup}
\label{section:hardware_setup}

\begin{figure}[ht]
  \centering
  \includegraphics[width=1.0\linewidth]{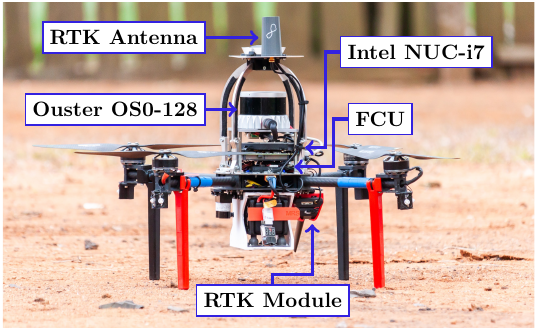}
  \caption{
    The MRS Drone quadrotor X500 platform that was used for dataset recording.
  }
  \label{fig:uav_hw}
\end{figure}

We have used the MRS Drone \cite{hert2023mrs} platform to record all datasets for evaluation.
The specific model we used is based on the Holybro X500 quadrotor frame with the Pixhawk 4 \ac{fcu} and Intel NUC10i7FNH onboard computer.
The \ac{lidar} scans for \ac{lo} are captured by the Ouster OS0-128 with integrated InvenSense ICM-20948 IMU.
The propulsion system consists of T-Motor P13$\times$4.4 carbon propellers mounted on T-Motor MN3510 KV700 motors controlled by Turnigy MultiStar BLheli32 51A \acp{esc}, which also provide the \ac{mas} measurements at \SI{80}{\hertz}.
For ground truth, we use the RTK solution from Emlid. 
The Emlid Reach M2 RTK module mounted on the drone is wirelessly linked to the Emlid Reach RS2 base station to provide GNSS positioning with a maximum error of \SI{7}{\milli\meter} in horizontal and \SI{14}{\milli\meter} in vertical direction.
All important components are highlighted in \autoref{fig:uav_hw}.

\subsection{Datasets}
\label{section:datasets}

We have recorded several outdoor flight sequences, each with a different kind of motion, to be able to identify what kind of motion correlates with high errors in which metrics.
The \textit{Hover} dataset is for stationary flight with the only motion being the takeoff and landing. 
Then, we have a dataset for each body axis of motion (\textit{Forward, Lateral, Vertical}), a \textit{Rectangle} with a constant heading of the \ac{uav}, and a \textit{Loop}, which is the longest dataset in both time and distance traveled, containing all motion combined, including heading changes.
See \autoref{tab:datasets} for dataset parameters.

The datasets contain \ac{mas} measurements, \ac{lidar} scans, \ac{imu} data, and \ac{rtk}-based ground truth.
Because the attitude cannot be obtained from \ac{rtk} using a single receiver, we use the orientation from the internal state estimation of the Pixhawk 4 \ac{fcu} \cite{meier2015px4}  as ground truth.
The last part of the ground truth, the linear velocity, was estimated by a Kalman filter from \cite{baca2021mrs}.

The datasets are open-sourced\footnote{\url{https://github.com/ctu-mrs/mas_datasets}} and available for free use.


    \begin{table}[ht]
        \setlength{\tabcolsep}{4pt}
        \centering
        \caption{\label{tab:datasets}
        The parameters of datasets used in the evaluation.
        The flight time is the time the \ac{uav} was in the air ($>$\SI{1}{\meter} from the ground after takeoff and $<$\SI{0.5}{\meter} during landing).
        The horizontal and vertical distances are the sums of relative displacements between consecutive ground truth messages and the velocities are the mean of ground truth velocities over the whole flight.
        }
        \centering
        \tablesize
        \begin{tabular}{lcccccc}
          \toprule
          & \textit{Hover} & \textit{Forward} & \textit{Lateral} & \textit{Vertical} & \textit{Rectangle} & \textit{Loop} \\
          \midrule
          \textbf{Flight time (\SI{}{\second})} & 52.21 & 107.01 & 89.83 & 46.18 & 109.64 & \textbf{197.17} \\
          \textbf{Hor. dist (\SI{}{\meter})} & 7.17 & 96.57 & 72.61 & 6.24 & 88.68 & \textbf{152.09} \\
          \textbf{Vert. dist (\SI{}{\meter})} & 14.66 & 10.05 & 10.30 & 35.00 & 17.13 & \textbf{43.74} \\
          \textbf{Hor. vel. (\SI{}{\meter\per\second})} & 0.09 & \textbf{0.87} & 0.73 & 0.08 & 0.76 & 0.73 \\
          \textbf{Vert. vel. (\SI{}{\meter\per\second})} & 0.25 & 0.08 & 0.08 & \textbf{0.72} & 0.10 & 0.19 \\
          \textbf{Yaw rate (\SI{}{\radian\per\second})} & 0.068 & 0.034 & 0.022 & 0.019 & 0.022 & \textbf{0.110} \\
          \bottomrule
        \end{tabular}
    \end{table}


\subsection{Noise evaluation}
\label{section:noise_evaluation}

\begin{figure}[t]
  \centering
  \adjincludegraphics[width=1.0\linewidth, trim={{0.00\width} {0.0\height} {0.0\width} {0.0\height}}, clip]{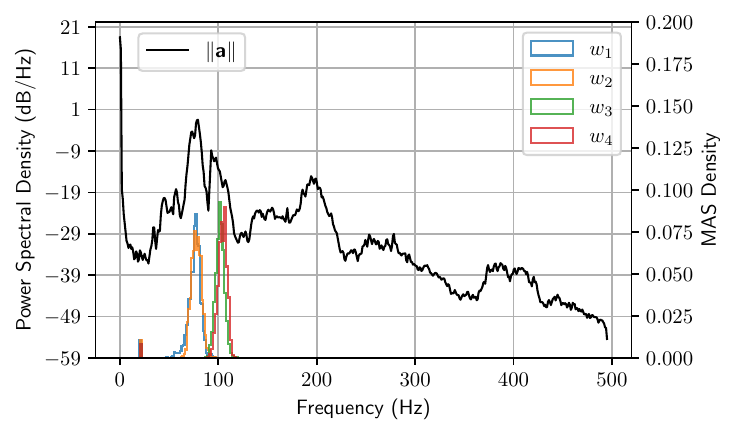}
  \caption{
    The power spectral density of the norm of the acceleration vector contains high peaks that correlate with the frequency of the spinning propellers.
    These vibrations are a major cause of errors in inertial localization methods.
    The data from this plot is taken from the \textit{Loop} dataset.
  }
  \label{fig:noise}
\end{figure}

As mentioned in~\autoref{section:inertial_methods}, the propeller-induced vibrations introduce significant error to the acceleration measurements of \ac{imu}, and thus degrade the performance of inertial localization methods.
This effect can be observed in~\autoref{fig:noise}, where we show the correlation of \ac{mas} and power spectral density of the norm of the acceleration vector.

\begin{figure}[t]
  \centering
  \adjincludegraphics[width=1.0\linewidth, trim={{0.00\width} {0.0\height} {0.0\width} {0.0\height}}, clip]{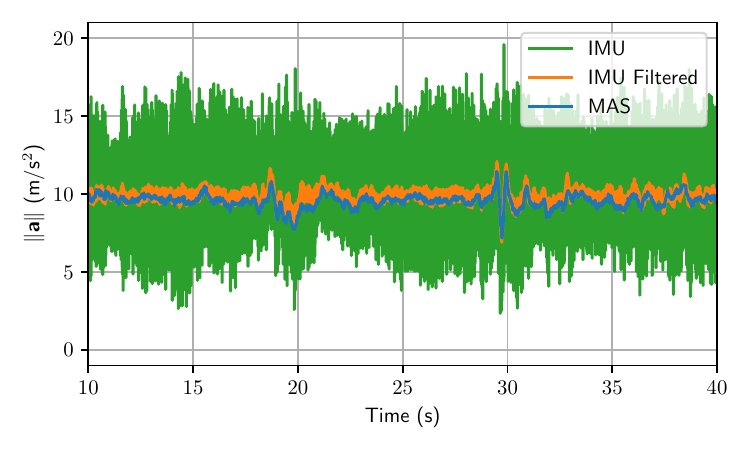}
  \caption{
    The propeller-induced high-frequency noise with amplitudes exceeding \SI{5}{\meter\per\second\squared} is superimposed on the actual acceleration norm. 
    To even be able to see the acceleration norm trajectory in the plot, we had to filter the signal with a low-pass filter.
    After filtering, the \ac{imu} trajectory has a similar shape to \ac{mas} preintegrated trajectory, although the noise is still higher.
    The data from this plot is taken from the \textit{Loop} dataset.
  }
  \label{fig:noise_compare}
\end{figure}

The amplitude of such vibrations is so large that it is not possible to visually compare the acceleration obtained from the \ac{mas} preintegration with the accelerations measured by \ac{imu}.
    To be able to even see the acceleration norm trajectory in the plot, we had to process the signal with a low-pass filter and notch filters at frequencies of propellers and their harmonics.
    After filtering, we can see in~\autoref{fig:noise_compare} that the \ac{mas} accelerations contain very little noise compared to even the filtered \ac{imu} accelerations.
    There is a noticeable varying offset between the trajectories, which can be caused by imprecise model parameters in the \ac{mas} and uncalibrated scale factor in the \ac{imu}.
    These are both accounted for in the form of bias terms.
    The mean of the \ac{imu} acceleration norm is $\SI{10.04}{\meter\per\second\squared}$ with a standard deviation of \SI{3.26}{\meter\per\second\squared} and for the \ac{mas} it is \SI{9.71}{\meter\per\second\squared} with a standard deviation \SI{0.30}{\meter\per\second\squared} on the \textit{Hover} dataset. 
    Except for takeoff and landing, the \ac{uav} was hovering during the flight, so the mean acceleration norm should be ideally \SI{9.81}{\meter\per\second\squared} with only a small standard deviation.

\subsection{Runtime evaluation}
\label{section:runtime_evaluation}

\begin{figure}[h]
  \centering
  \adjincludegraphics[width=1.0\linewidth, trim={{0.00\width} {0.0\height} {0.0\width} {0.0\height}}, clip]{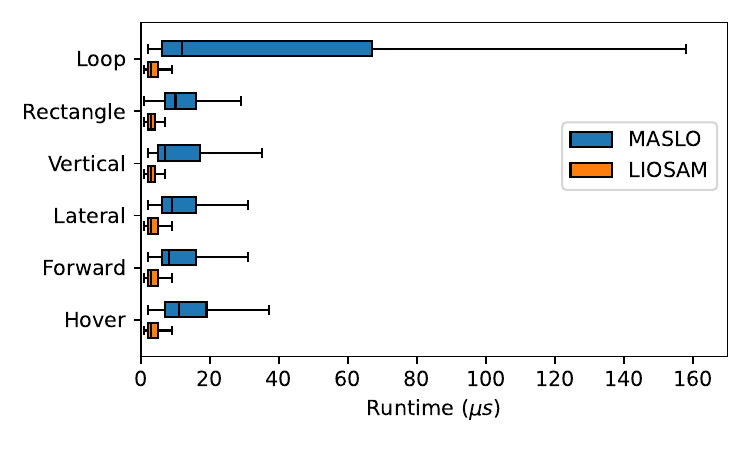}
  \caption{
    Histogram of \ac{mas} preintegration times in \ac{maslo} and \ac{imu} preintegration times in \ac{liosam}.
    The outliers are not shown in this boxplot to keep it clear.
  }
  \label{fig:runtime}
\end{figure}

The \ac{mas} preintegration must be sufficiently fast to be able to process high-frequency \ac{mas} measurements.
The median preintegration runtime over all datasets is \SI{10}{\micro\second}, with \autoref{fig:runtime} showing the distribution of runtimes in individual datasets.
Despite being higher than \SI{3}{\micro\second} of \ac{imu} preintegration, the runtime is still orders of magnitude faster than necessary for real-time operation. 
The runtime still theoretically allows the preintegration to run at up to \SI{100}{\kilo\hertz}.

In our datasets, the \ac{mas} is reported at \SI{80}{\hertz}, so the theoretical maximum preintegration frequency is three orders of magnitude higher than the frequency of \ac{mas} measurements.

\begin{figure}[b]
  \centering
  \adjincludegraphics[width=\linewidth, trim={{0.00\width} {0.0\height} {0.0\width} {0.0\height}}, clip]{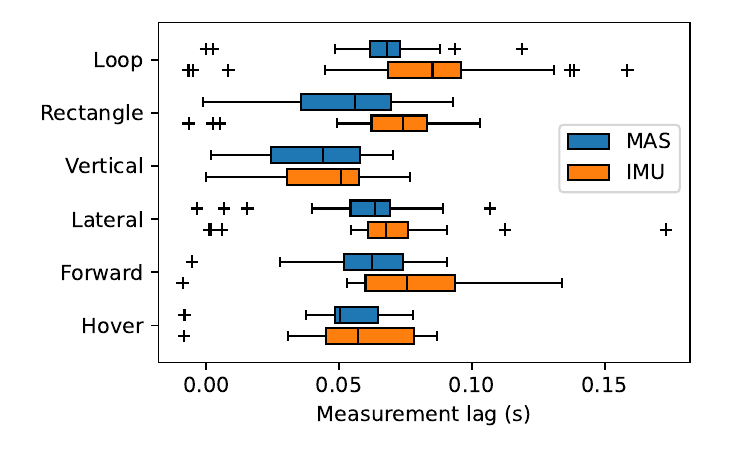}
  \caption{
    Boxplot of the lag between the measured accelerations and the target control accelerations.
    The values were obtained by evaluating \SI{5}{\second} segments of the whole trajectory.
  }
  \label{fig:measurement_lag}
\end{figure}



\begin{figure*}
    \newcommand{\bluedot}{\raisebox{0pt}{\tikz{\draw[tab_blue,fill=tab_blue] (0,0) circle (2.0pt);}}}
    \newcommand{\greenplus}{\hspace{-3pt}\raisebox{-2pt}{\tikz{\node[text=tab_green]{\textbf{+}};}}\hspace{-3pt}}
    \newcommand{\orangecross}{\hspace{-3pt}\raisebox{-4pt}{\tikz{\node[text=tab_orange]{{$\bm{\times}$}};}}\hspace{-3pt}}
    \newcommand{\redsquare}{\hspace{-2pt}\raisebox{-3pt}{\tikz{\node[text=tab_red]{{$\blacksquare$}};}}\hspace{-2pt}}
    \newcommand{\imheight}{10.8em}
    \newcommand{\xcap}{1.3em}
    \newcommand{\ycap}{-10.5em}
    \newcommand{\fillopa}{0.3}
    \centering
    \adjincludegraphics[width=\linewidth, trim={{0.0\width} {0.0\height} {0.0\width} {0.0\height}}, clip]{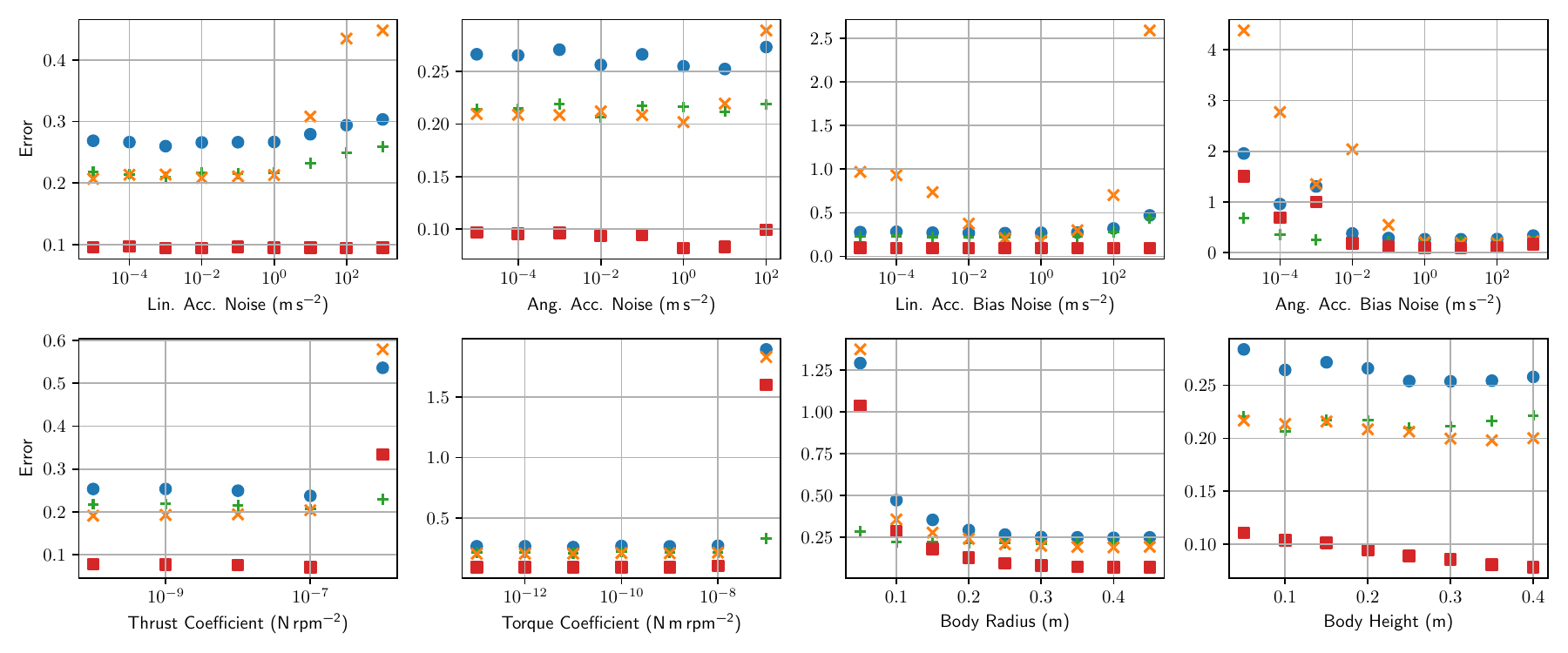}
  \caption{
    The effect of changing \ac{mas} parameter values on the trajectory error evaluated using the \ac{ape} (\protect\bluedot), \ac{ate} (\protect\greenplus), \ac{ave} (\protect\orangecross), \ac{are}\protect\linebreak (\protect\redsquare) metrics.     
    Note that only the Body Radius and Body Height plots have linear x-axis.
    All other plots have logarithmic x-axis.
    The data for this plot is obtained by running the \ac{maslo} algorithm on the \textit{Loop} dataset.
  }
  \label{fig:param_values}
\end{figure*}
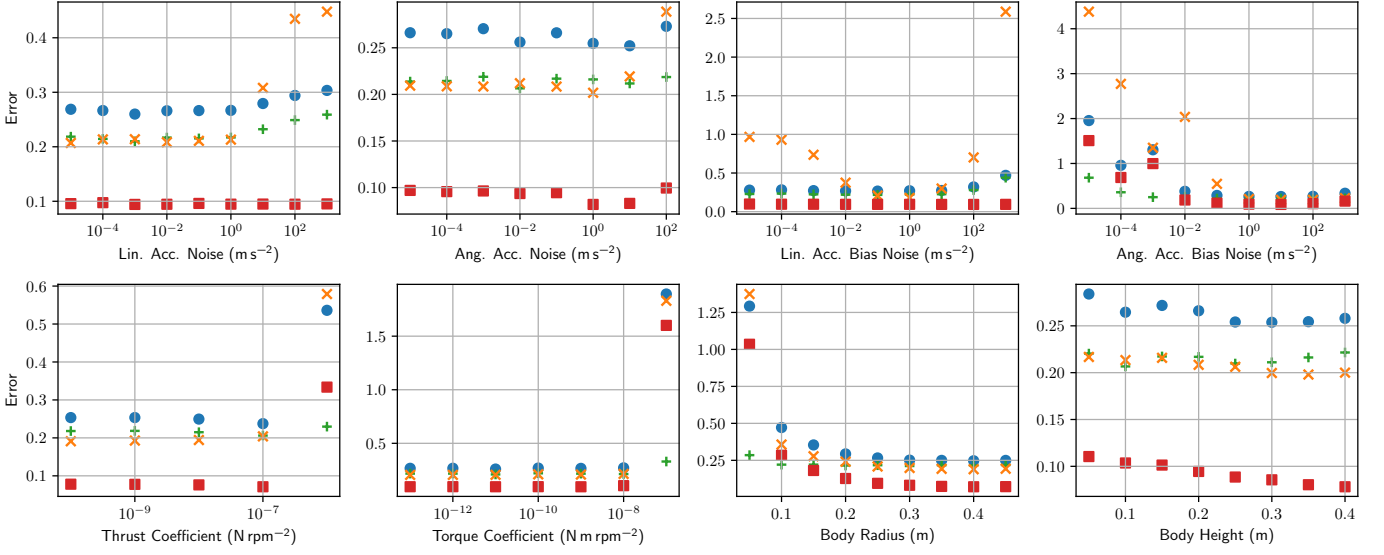


\subsection{Measurement lag}
\label{section:lag_evaluation}

For precise control and especially for agile flights the crucial property of any measurements fused in state estimation pipelines is the measurement lag, i.e., the time delay of the measured value w.r.t. the true value.
Ground truth measurements used in evaluation of localization methods for \acp{uav} typically do not contain the accelerations values, and thus we evaluate the delay w.r.t. the desired accelerations calculated by the controller.
It is not possible to obtain the absolute value of the measurement lag as the measured delay is the sum of the actual measurement lag and the delay coming from the physical system dynamic response that is mostly affected by the transient response of the propellers.
Nevertheless, this unknown system delay is the same for both \ac{imu} and \ac{mas} preintegration, and thus the relative comparison of the measurement lag of both methods is viable.

The measurement delay is obtained by shifting the measurement in time to align the measured and desired accelerations.
However, finding the correct alignment is sometimes not feasible due to noise and control errors.
To reduce the effect of the incorrectly aligned parts of the trajectory, we have divided it into sections of \SI{5}{\second}, aligned the sections individually and calculated the median measurement lag over all thus evaluated sections.
The distribution of the measurement lag calculated in this way is visualized in~\autoref{fig:measurement_lag}, where outliers attain even negative values, which points to failed alignment.

The median measurement lag obtained for the filtered \ac{imu} acceleration is \SI{0.073}{\second} and \SI{0.063}{\second} for the preintegrated \ac{mas} acceleration, resulting in \SI{14}{\percent} lower measurement lag of the \ac{mas} acceleration.
The higher lag of the \ac{imu} is most likely caused by the low-pass filter, which is used to reduce the amplitude of high-frequency propeller noise.


\subsection{Parameter values}
\label{section:param_values}

    \begin{table}
        \setlength{\tabcolsep}{4pt}
        \centering
        \caption{\label{tab:parameter_values}
        The values of parameters used in the evaluation (Default), together with the best found value, and range of values for reasonable performance.
        }
        \centering
        \tablesize
        \begin{tabular}{lccc}
          \toprule
          & \textbf{Default} & \textbf{Best} & \textbf{Valid range} \\
          \midrule
          \textbf{Lin. Acc. Noise $\eta_a$ (\SI{}{\meter\per\second\squared})} & 0.1 & 0.001 & 10$^{-5}$--10$^0$  \\
          \textbf{Ang. Acc. Noise $\eta_{\alpha}$ (\SI{}{\radian\per\second\squared})} & 1 & 10 & 10$^{-5}$--10$^1$  \\
          \textbf{Lin. Acc. Noise Bias $\nu_{a}$ (\SI{}{\meter\per\second\squared})} & 0.1 & 0.1 & 10$^{-2}$--10$^1$  \\
          \textbf{Ang. Acc. Noise Bias $\nu_{\alpha}$ (\SI{}{\radian\per\second\squared})} & 1 & 10 & 10$^{-1}$--10$^{-3}$  \\
          \textbf{Thrust Coefficient $c_f$ (\SI{}{\newton\per\rpm\squared})} & 2.6$\cdot$10$^{-7}$ & 10$^{-7}$ & 10$^{-10}$--10$^{-7}$  \\
          \textbf{Torque Coefficient $c_d$ (\SI{}{\newton\meter\per\rpm\squared})} & 2.6$\cdot$10$^{-9}$ & 2.6$\cdot$10$^{-11}$ & 10$^{-13}$--10$^{-8}$  \\
          \textbf{Body Radius $r$ (\SI{}{\meter})} & 0.25 & 0.4 & 0.2--0.45  \\
          \textbf{Body Height $h$ (\SI{}{\meter})} & 0.2 & 0.3 & 0.2--0.45  \\
          \bottomrule
        \end{tabular}
    \end{table}
As any model-based algorithm, the accuracy of obtaining accelerations from \ac{mas} is dependent on the correct values of the parameters of the model. 
On its own, the \ac{mas} preintegration would be very sensitive to the parameter values.
Finding the correct parameter values would be too strenuous without some margin for error.
However, thanks to the bias term included in \ac{maslo}, the state estimation performance is robust to inaccurate parameter guesses to some degree.

Here we present a parameter value study to establish the sensitivity to correct parameter values.
From~\autoref{fig:param_values} we can see that \ac{maslo} is not very sensitive to the parameter values, and even an educated guess without further tuning of the values is sufficient for reasonable performance.
\autoref{tab:parameter_values} lists default parameter values that were used for all experiments, along with the best value found (not necessarily the same as the default) and the range of values that provides an acceptable performance.

The mass of the \ac{uav} was measured and fixed to $m=$\SI{3.2}{\kilogram}.
The body radius and height are used in the approximation of the \ac{uav} inertia matrix from~\autoref{eq:rot_dynamics}:
\begin{equation}
  \mathbf{M}=  
  \left[
\begin{smallmatrix}
  I_x & 0 & 0 \\
  0  & I_y & 0 \\
  0 & 0 & I_z 
\end{smallmatrix}
  \right],
\end{equation}
where 
\begin{equation}
  I_x = I_y = \frac{1}{12}m\left(3r^2+h^2\right), \quad
  I_z = \frac{1}{2}m r^2.
\end{equation}

Note that this study is not an exhaustive grid-search of the best parameter combination and each parameter was evaluated individually with the other parameters having the default value.
The obtained values are naturally valid only for the specific \ac{uav} platform used to record the datasets.


\subsection{Outdoor flights evaluation}
\label{section:outdoor_evaluation}

In this section, we will evaluate the performance of the proposed \ac{mas} factor.
As far as we know, there is no other localization method using the \ac{mas} measurements for pose estimation that we could compare our method to.
Hence, we will compare \ac{maslo}, to \ac{liosam}, which are essentially the same algorithms with the key difference being the preintegration of \ac{mas} and \ac{imu} measurements, respectively.

\subsubsection{Qualitative analysis}
\label{section:qualitative_analysis}
\begin{figure*}
  \newcommand{\blueline}{\raisebox{2pt}{\tikz{\draw[tab_blue,solid,line width = 1.0pt](0,0) -- (5mm,0);}}}
  \newcommand{\orangeline}{\raisebox{2pt}{\tikz{\draw[tab_orange,solid,line width = 1.0pt](0,0) -- (5mm,0);}}}
  \newcommand{\blackline}{\raisebox{2pt}{\tikz{\draw[black,solid,line width = 1.0pt](0,0) -- (5mm,0);}}}
  \centering
  \adjincludegraphics[width=1.0\linewidth, trim={{0.00\width} {0.0\height} {0.0\width} {0.0\height}}, clip]{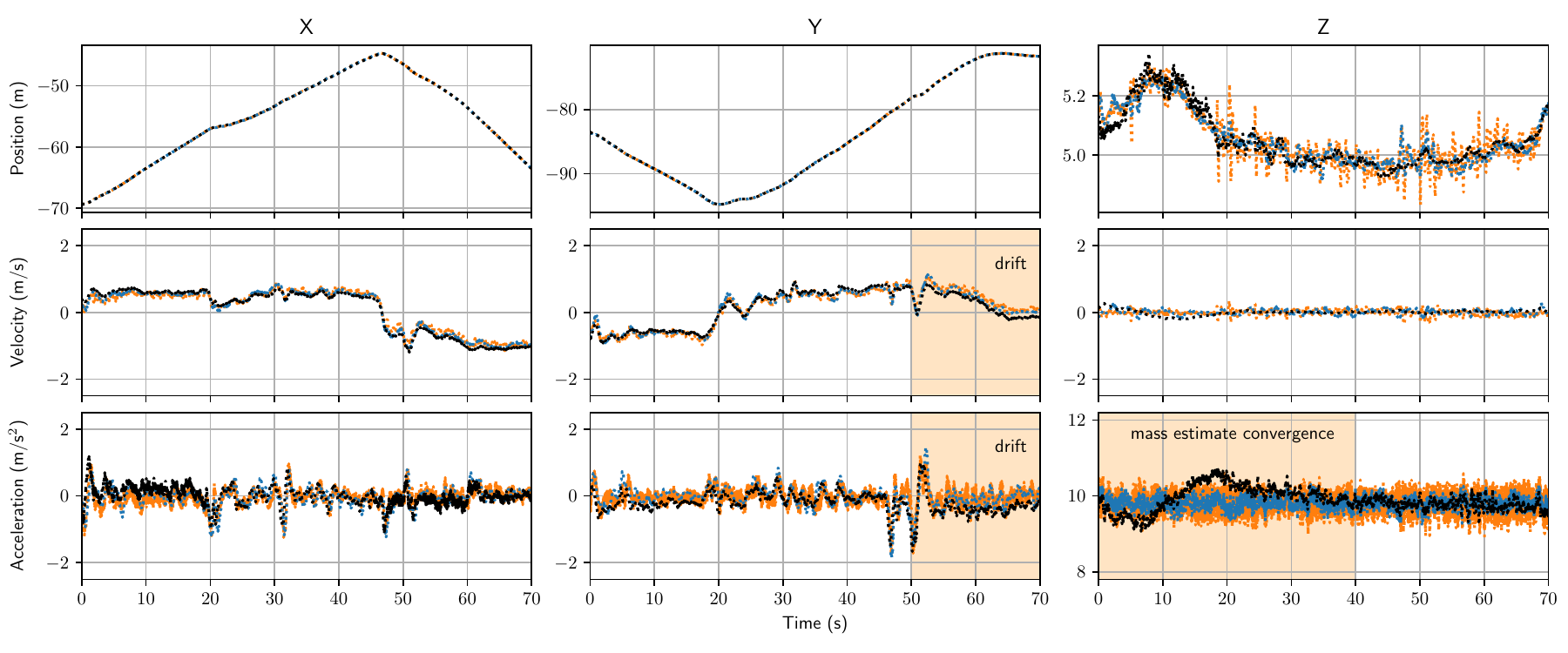}
  \caption{
    The position, velocity, and acceleration (all in world frame) estimated by \ac{maslo} (\protect\blueline) and \ac{liosam} (\protect\orangeline).
    Position and velocity plots also contain the ground truth values (\protect\blackline).
    A noticeable drift in y-axis velocity and acceleration is highlighted.
    As ground truth acceleration is not available in the datasets, the acceleration plots shows the desired control acceleration.
    The highlighted part has too large control inputs due to the not-yet-converged mass estimator.
    The data for this plot is obtained by running the algorithms on the \textit{Rectangle} dataset.
  }
  \label{fig:states}
\end{figure*}

\begin{figure*}
  \newcommand{\blueline}{\raisebox{2pt}{\tikz{\draw[tab_blue,solid,line width = 1.0pt](0,0) -- (5mm,0);}}}
  \newcommand{\orangeline}{\raisebox{2pt}{\tikz{\draw[tab_orange,solid,line width = 1.0pt](0,0) -- (5mm,0);}}}
  \newcommand{\blackline}{\raisebox{2pt}{\tikz{\draw[black,solid,line width = 1.0pt](0,0) -- (5mm,0);}}}
  \centering
  \adjincludegraphics[width=1.0\linewidth, trim={{0.00\width} {0.0\height} {0.0\width} {0.0\height}}, clip]{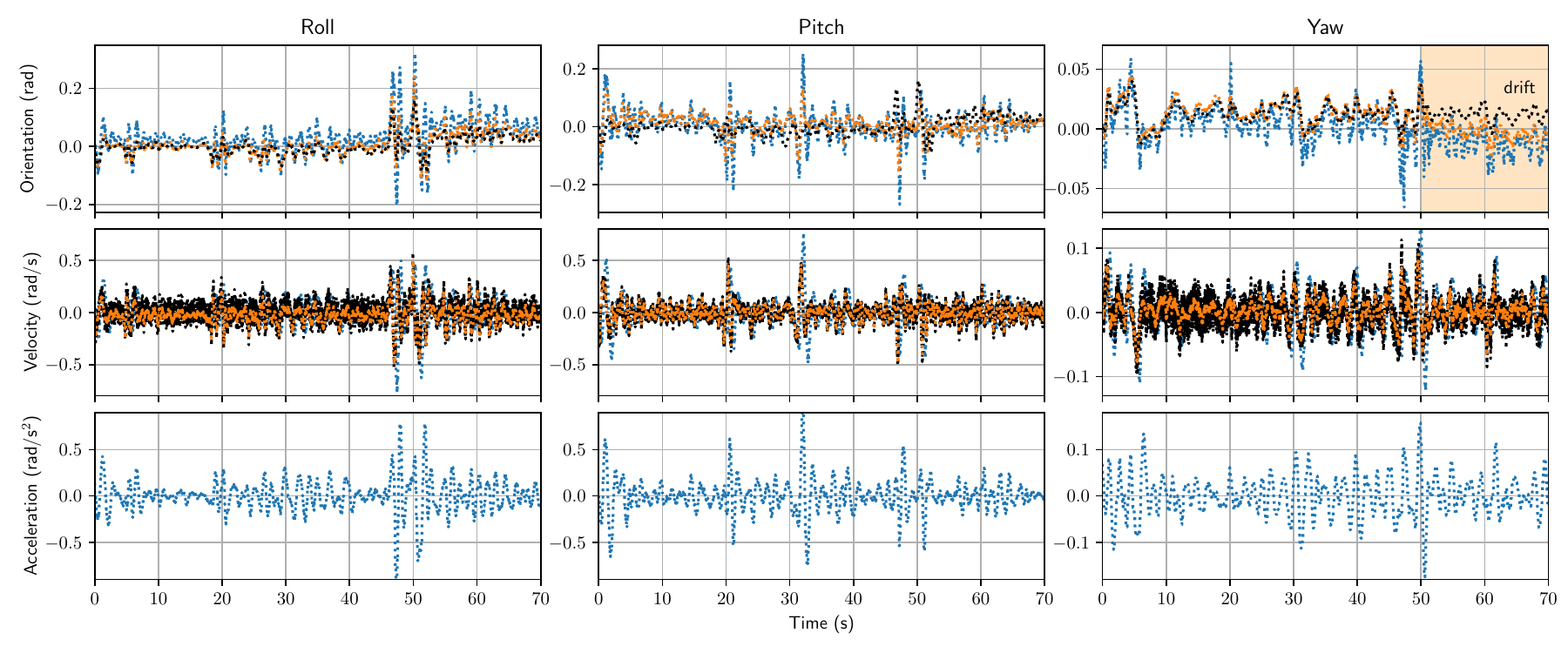}
  \caption{
    The orientation, angular velocity, and angular acceleration estimated by \ac{maslo} (\protect\blueline) and \ac{liosam} (\protect\orangeline).
    Orientation and velocity plots also contain the ground truth values (\protect\blackline).
    A noticeable drift in the yaw angle is highlighted.
    The acceleration plots shows only the \ac{mas} acceleration estimate, as \ac{liosam} does not provide angular accelerations, and neither there is any ground truth angular acceleration in the datasets. 
    The data for this plot is obtained by running the algorithms on the \textit{Rectangle} dataset.
  }
  \label{fig:orientation}
\end{figure*}

We have plotted the translational part of the \ac{uav} state in~\autoref{fig:states} and the rotational part in~\autoref{fig:orientation} to be able to better visualize the differences between the algorithms.
The plots were obtained by running the algorithms on the \textit{Rectangle} dataset and then the estimated trajectories were aligned to ground truth to show them in the same coordinate frame.
The translational estimation of \ac{liosam} seems to be subjectively more noisy, which can be seen mostly in the z-axis position and acceleration plots.
Both algorithms also exhibit a noticeable drift close to the end of the flight (after \SI{50}{\second}) in the y-axis velocity and acceleration.
The control acceleration in the first half of the z-axis plot has a large bias caused by the inaccurately estimated mass in the controller.
The desired control inputs were thus higher values than the physical model required.
Around half of the flight the mass estimate converged to the correct value.

The orientation from \ac{maslo} is more noisy,  both algorithms are biased in the yaw towards the end of the flight and a smaller bias in yaw is also noticeable for \ac{maslo} since the beginning of flight.
The reference trajectory for the angular velocity is reported by \ac{fcu} \ac{imu} and as such it is not really a ground truth but it shows that the angular velocities are correctly (although with larger noise) estimated by \ac{maslo}, which does not use any \ac{imu}.
The \ac{liosam} angular velocity is also less noisy than the reference acceleration because \ac{liosam} was run on the filtered \ac{imu} data, in contrast to the \ac{fcu} \ac{imu} which was not filtered in this plot.
The only angular acceleration trajectory is available from \ac{maslo}.
Neither \ac{liosam} nor ground truth have angular acceleration available so we plot only \ac{maslo} to see that the trajectory attains reasonable values and is without any noticeable bias and excessive noise.

\begin{figure}[tb]
    \newcommand{\imheight}{10.8em}
    \newcommand{\xcap}{1.3em}
    \newcommand{\ycap}{-10.5em}
    \newcommand{\fillopa}{0.3}
  \centering
    \begin{tikzpicture}
      \node[anchor=north west,inner sep=0] (b) at (0,0) {\adjincludegraphics[width=0.49\linewidth, trim={{0.06\width} {0.0\height} {0.2\width} {0.2\height}}, clip]{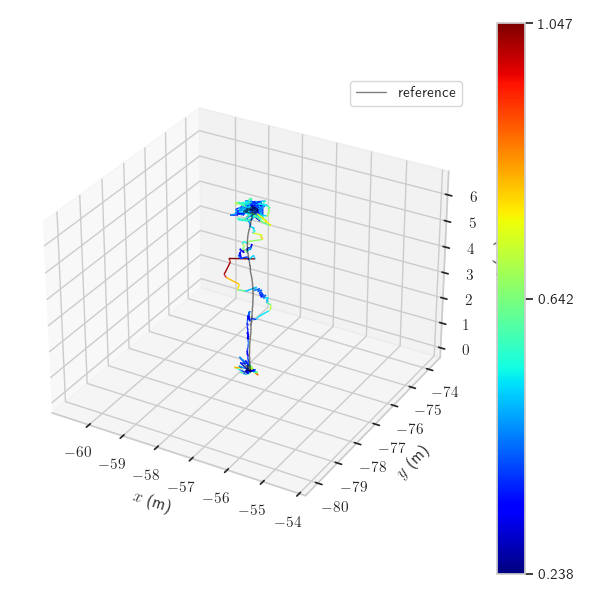}};%
    \begin{scope}[x={(b.south east)},y={(b.north west)}]
      \node[fill=white, fill opacity=\fillopa, text=black, text opacity=1.0] at (\xcap, \ycap) {\textbf{(a)}};
      \end{scope}
    \end{tikzpicture}
    \vspace{-2em}
    \begin{tikzpicture}
      \node[anchor=north west,inner sep=0] (b) at (0,0) {\adjincludegraphics[width=0.49\linewidth, trim={{0.06\width} {0.0\height} {0.2\width} {0.2\height}}, clip]{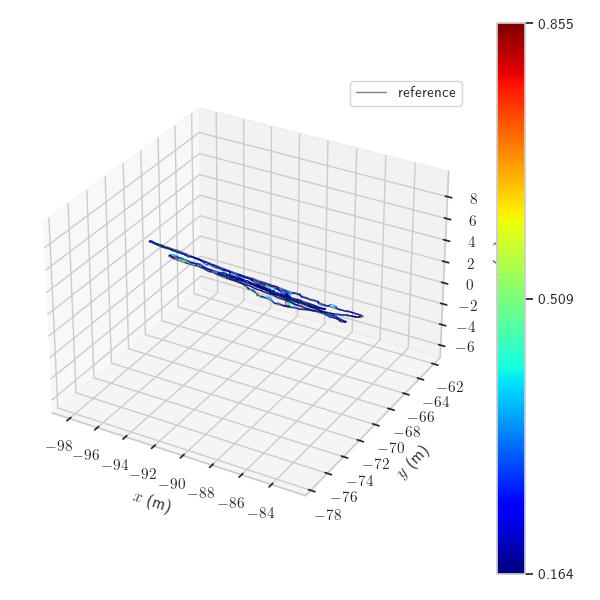}};%
    \begin{scope}[x={(b.south east)},y={(b.north west)}]
      \node[fill=white, fill opacity=\fillopa, text=black, text opacity=1.0] at (\xcap, \ycap) {\textbf{(b)}};
      \end{scope}
    \end{tikzpicture}
    \vspace{-2em}
    \begin{tikzpicture}
      \node[anchor=north west,inner sep=0] (b) at (0,0) {\adjincludegraphics[width=0.49\linewidth, trim={{0.06\width} {0.0\height} {0.2\width} {0.2\height}}, clip]{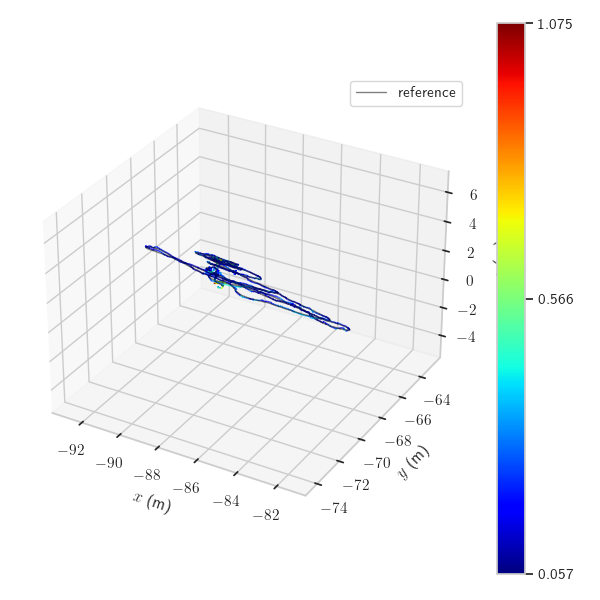}};%
    \begin{scope}[x={(b.south east)},y={(b.north west)}]
      \node[fill=white, fill opacity=\fillopa, text=black, text opacity=1.0] at (\xcap, \ycap) {\textbf{(c)}};
      \end{scope}
    \end{tikzpicture}%
    \begin{tikzpicture}
      \node[anchor=north west,inner sep=0] (b) at (0,0) {\adjincludegraphics[width=0.49\linewidth, trim={{0.06\width} {0.0\height} {0.2\width} {0.2\height}}, clip]{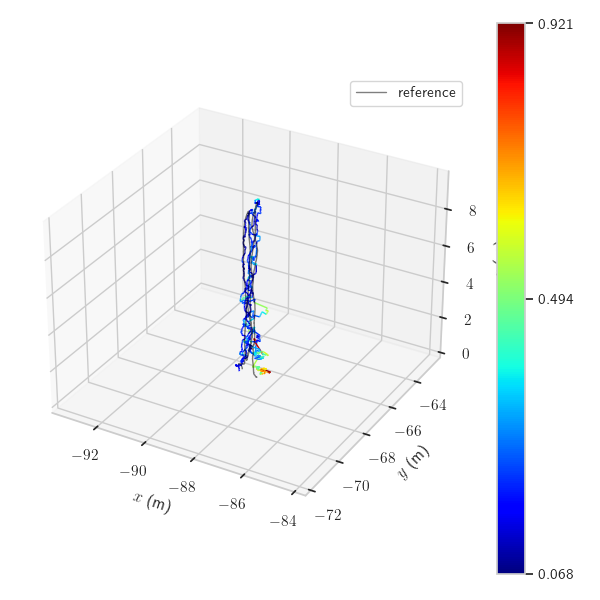}};%
    \begin{scope}[x={(b.south east)},y={(b.north west)}]
      \node[fill=white, fill opacity=\fillopa, text=black, text opacity=1.0] at (\xcap, \ycap) {\textbf{(d)}};
      \end{scope}
    \end{tikzpicture}
    \begin{tikzpicture}
      \node[anchor=north west,inner sep=0] (b) at (0,0) {\adjincludegraphics[width=0.49\linewidth, trim={{0.06\width} {0.0\height} {0.19\width} {0.2\height}}, clip]{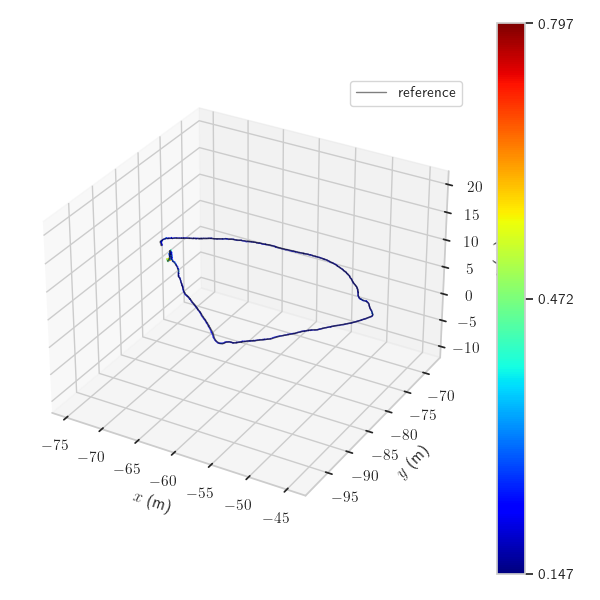}};%
    \begin{scope}[x={(b.south east)},y={(b.north west)}]
      \node[fill=white, fill opacity=\fillopa, text=black, text opacity=1.0] at (\xcap, \ycap) {\textbf{(e)}};
      \end{scope}
    \end{tikzpicture}%
    \begin{tikzpicture}
      \node[anchor=north west,inner sep=0] (b) at (0,0) {\adjincludegraphics[width=0.49\linewidth, trim={{0.06\width} {0.0\height} {0.19\width} {0.2\height}}, clip]{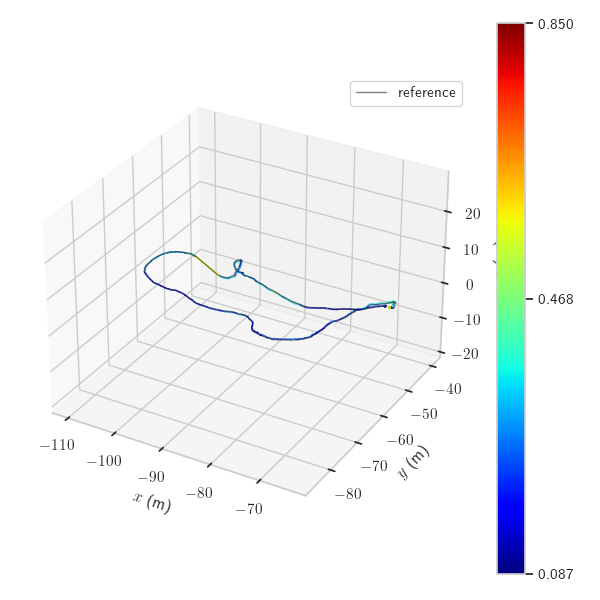}};%
    \begin{scope}[x={(b.south east)},y={(b.north west)}]
      \node[fill=white, fill opacity=\fillopa, text=black, text opacity=1.0] at (\xcap, \ycap) {\textbf{(f)}};
      \end{scope}
    \end{tikzpicture}%
  \caption{
    The trajectories estimated by \ac{maslo} on the recorded datasets (a): \textit{Hover}, (b): \textit{Forward}, (c): \textit{Lateral}, (d): \textit{Vertical}, (e): \textit{Rectangle}, (f): \textit{Loop}).
    The color of the trajectory indicates the mean \ac{ate} (lowest error in blue, highest error in red) w.r.t. the \ac{rtk} ground truth (gray).
  }
  \label{fig:trajectories}
\end{figure}
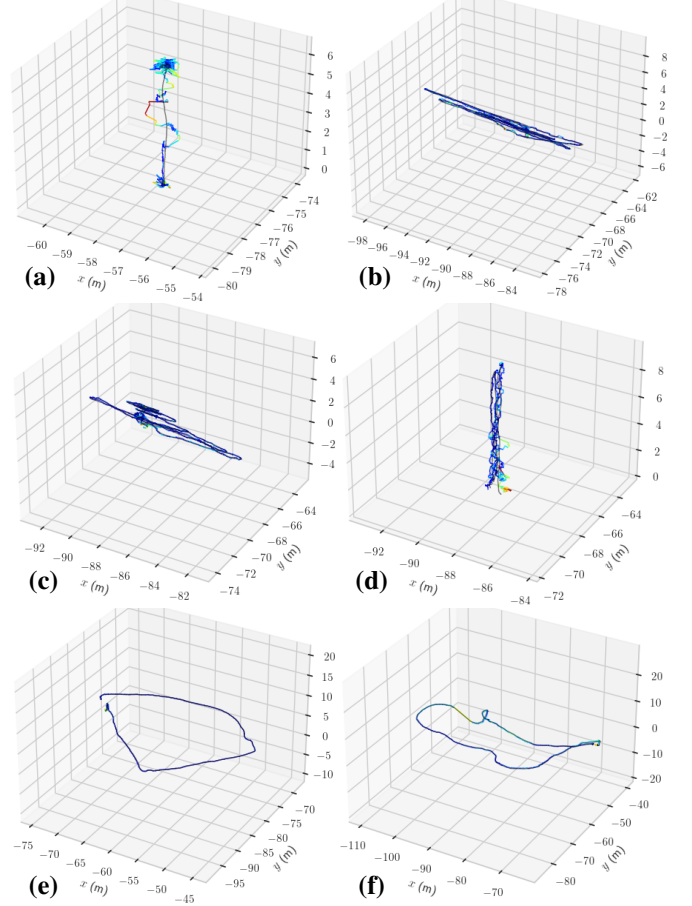


\subsubsection{Metrics}
\label{section:metrics}

For quantitative evaluation of the localization algorithms, we use the most common metric: the \ac{ape}, which outputs a single error value calculated from the difference of estimated pose w.r.t. the corresponding (closest timestamp) ground truth pose \cite{zhang2018tutorial}.
Since the estimated and ground truth trajectories are in different coordinate frames, they are first aligned before calculating the error, which also prevents rotation errors at the start of the trajectory from having a disproportionately higher influence on the metric than rotation errors at the end of the trajectory.
The metric is split into a translation part (\ac{ate}) and rotation part (\ac{are}) for better insight into the source of error and to prevent the mixing of different units.
We based our evaluation on the evo framework \cite{grupp2017evo}, which we modified by further splitting the rotation part into errors in individual axes to see if certain motions induce higher errors.
We added another metric, the linear velocity error (\ac{ave}), which is often not evaluated in other literature. However, depending on the controller, it may be critical for the stabilization of the \ac{uav}.


\subsubsection{Quantitative evaluation}
\label{section:results}
\begin{figure}[t]
  \centering
  \adjincludegraphics[width=0.49\linewidth, trim={{0.00\width} {0.0\height} {0.0\width} {0.0\height}}, clip]{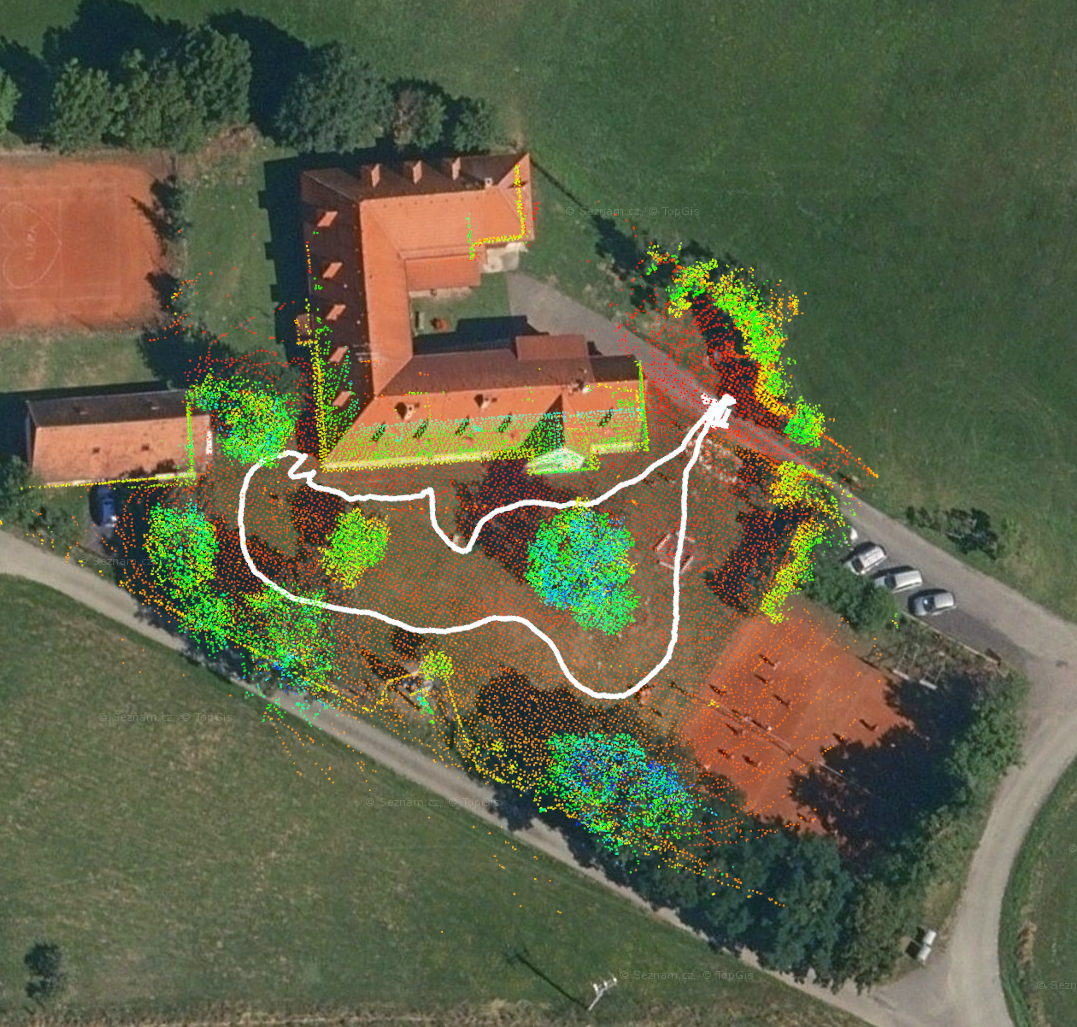}
  \adjincludegraphics[width=0.49\linewidth, trim={{0.00\width} {0.0\height} {0.0\width} {0.0\height}}, clip]{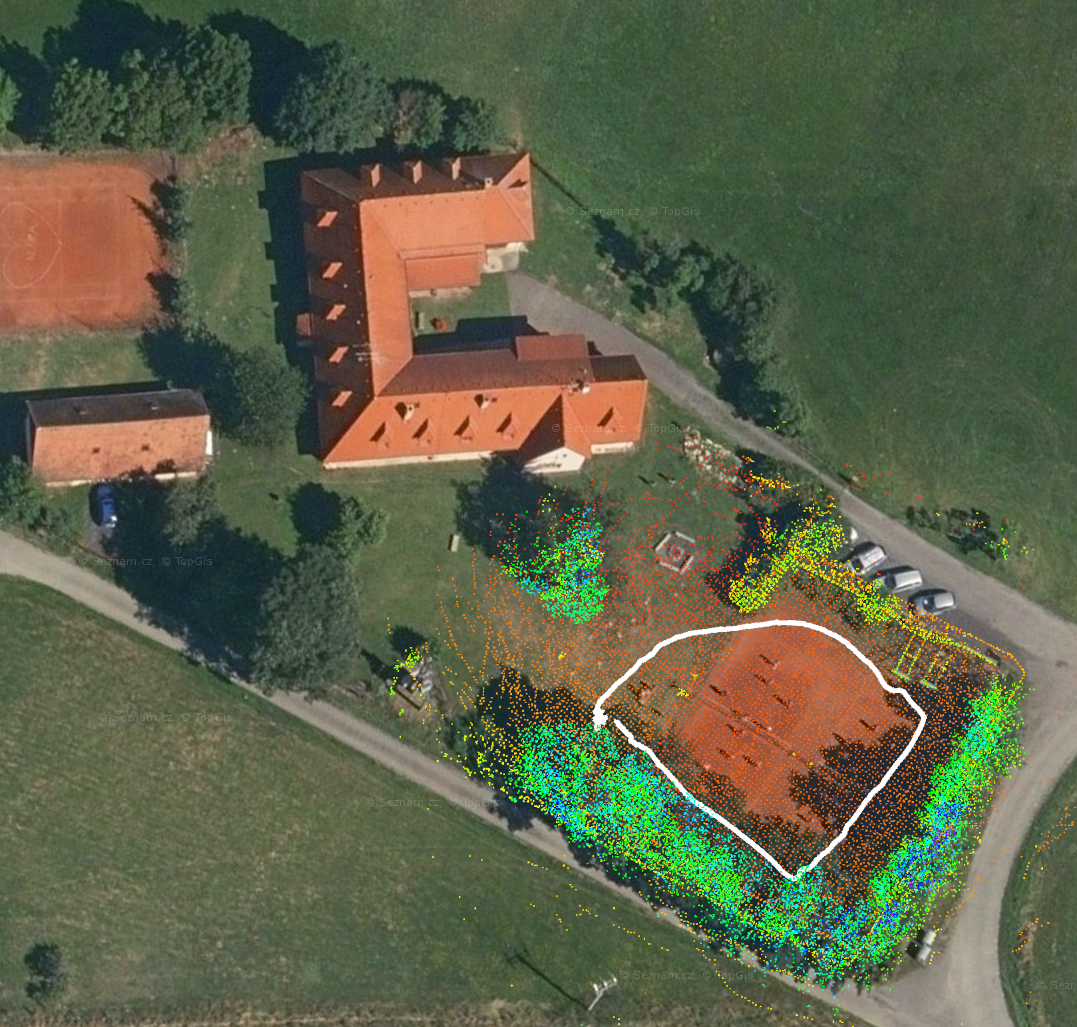}
  \caption{
    The satellite image of the dataset location with \ac{maslo} map overlay from the \textit{Loop} dataset on the left and \textit{Rectangle} dataset on the right.
    The estimated trajectory is shown in white.
  }
  \label{fig:maps}
\end{figure}

Looking at the errors in~\autoref{tab:evaluation}, we find \ac{maslo} outperforming \ac{liosam} in all datasets, except for \textit{Hover} in the translation metric.
\ac{maslo} reaches 28\% lower \ac{ate} than \ac{liosam} when we average the error over all datasets.
In linear velocity evaluation, \ac{maslo} outperforms \ac{liosam} in all datasets, including the \textit{Hover} dataset, which can be seen in the \ac{ave} metric that is 65\% lower than in the case of \ac{liosam}.
On the other hand, the rotation estimation is better in \ac{liosam} in all datasets with \ac{are} of \ac{maslo} being 20\% higher.
The worse performance in rotation is not surprising considering the additional integration term in the \ac{maslo} rotational model when compared to the \ac{liosam} model.
It is worth mentioning that \ac{maslo} has the worst results in the \textit{Hover} dataset.
The \textit{Hover} dataset differs from the other datasets in the lack of motion, therefore the worse result on this dataset could point to a diverging estimate due to insufficient excitation.
We further decompose the \ac{are} metric into errors in the roll, pitch, and yaw rotations in~\autoref{tab:rotation_error}, which shows that except the \textit{Hover} dataset, the errors in the yaw rotations are lower than in roll and pitch.
The maps and trajectories from \textit{Loop} and \textit{Rectangle} datasets are shown in~\autoref{fig:trajectories} and~\autoref{fig:maps} shows the distribution of error along the trajectories in 3D space.

    \begin{table}[t]
        \setlength{\tabcolsep}{0pt}
        \centering
        \caption{\label{tab:evaluation}
        The mean error $\mu$ and its standard deviation $\sigma$ of translation, rotation, and linear velocity estimated by \ac{maslo} and \ac{liosam} w.r.t. ground truth.
        The best result out of the two algorithms for each metric and dataset is highlighted in bold.
        }
        \centering
        \tablesize
        \newcommand{\colwidth}{2.6em}
        \begin{tabular}{lC{\colwidth}C{\colwidth}C{\colwidth}C{\colwidth}C{\colwidth}C{\colwidth}C{\colwidth}C{\colwidth}C{\colwidth}C{\colwidth}C{\colwidth}C{\colwidth}C{\colwidth}C{\colwidth}C{\colwidth}C{\colwidth}}
          \toprule
          & \multicolumn{4}{c}{\tablehdg{Translation (\SI{}{\meter})}} & \multicolumn{4}{c}{\tablehdg{Rotation (\SI{}{\radian})}}  & \multicolumn{4}{c}{\tablehdg{Linear Vel. (\SI{}{\meter\per\second})}}\\
          & \multicolumn{2}{c}{\tablehdg{LIO-SAM}} & \multicolumn{2}{c}{\tablehdg{MAS-LO}} & 
           \multicolumn{2}{c}{\tablehdg{LIO-SAM}} & \multicolumn{2}{c}{\tablehdg{MAS-LO}} & 
           \multicolumn{2}{c}{\tablehdg{LIO-SAM}} & \multicolumn{2}{c}{\tablehdg{MAS-LO}}\\
          & \tablehdg{$\mu$} & \tablehdg{$\sigma$} & \tablehdg{$\mu$} & \tablehdg{$\sigma$} & \tablehdg{$\mu$} & \tablehdg{$\sigma$} & \tablehdg{$\mu$} & \tablehdg{$\sigma$} & \tablehdg{$\mu$} & \tablehdg{$\sigma$} & \tablehdg{$\mu$} & \tablehdg{$\sigma$} \\
          \midrule
          \textit{Hover}     & \textbf{0.18} & \textbf{0.12} & 0.24 & 0.13 & \textbf{0.19} & \textbf{0.05} & 0.23 & 0.06 & 1.09 & 0.85 & \textbf{0.50} & \textbf{0.48} \\
          \textit{Forward}   & 0.16 & 0.08 & \textbf{0.15} & \textbf{0.07} & \textbf{0.14}  & \textbf{0.02} & 0.17  & 0.05 & 0.56 & 0.40 & \textbf{0.26} & \textbf{0.29} \\
          \textit{Lateral}   & 0.32 & 0.27 & \textbf{0.13} & \textbf{0.07} & \textbf{0.07}  & \textbf{0.02} & 0.11  & 0.07 & 0.70 & 0.51 & \textbf{0.29} & \textbf{0.37} \\
          \textit{Vertical}  & 0.34 & 0.31 & \textbf{0.20} & \textbf{0.12} & \textbf{0.08}  & \textbf{0.02} & 0.09  & 0.06 & 0.79 & 0.65 & \textbf{0.33} & \textbf{0.41} \\
          \textit{Rectangle} & 0.25 & 0.22 & \textbf{0.09} & \textbf{0.05} & \textbf{0.12}  & \textbf{0.03} & 0.13  & 0.04 & 1.01 & 0.85 & \textbf{0.23} & \textbf{0.33} \\
          \textit{Loop}      & 0.33 & 0.26 & \textbf{0.23} & \textbf{0.08} & \textbf{0.06}  & \textbf{0.03} & 0.08  & 0.05 & 1.35 & 0.71 & \textbf{0.22} & \textbf{0.23} \\
          \bottomrule
        \end{tabular}
    \end{table}

    \begin{table}[t]
        \setlength{\tabcolsep}{0pt}
        \centering
        \caption{\label{tab:rotation_error}
        The mean error $\mu$ and its standard deviation $\sigma$ of roll, pitch, and yaw angles estimated by \ac{maslo} and \ac{liosam} w.r.t. ground truth.
        The best result out of the two algorithms for each metric and dataset is highlighted in bold.
        }
        \centering
        \tablesize
        \newcommand{\colwidth}{2.6em}
        \begin{tabular}{lC{\colwidth}C{\colwidth}C{\colwidth}C{\colwidth}C{\colwidth}C{\colwidth}C{\colwidth}C{\colwidth}C{\colwidth}C{\colwidth}C{\colwidth}C{\colwidth}C{\colwidth}C{\colwidth}C{\colwidth}C{\colwidth}}
          \toprule
          & \multicolumn{4}{c}{\tablehdg{Roll (\SI{}{\radian})}} & \multicolumn{4}{c}{\tablehdg{Pitch (\SI{}{\radian})}} & \multicolumn{4}{c}{\tablehdg{Yaw (\SI{}{\radian})}} \\
          & \multicolumn{2}{c}{\tablehdg{LIO-SAM}} & \multicolumn{2}{c}{\tablehdg{MAS-LO}} & 
           \multicolumn{2}{c}{\tablehdg{LIO-SAM}} & \multicolumn{2}{c}{\tablehdg{MAS-LO}} & 
           \multicolumn{2}{c}{\tablehdg{LIO-SAM}} & \multicolumn{2}{c}{\tablehdg{MAS-LO}} \\ 
          & \tablehdg{$\mu$} & \tablehdg{$\sigma$} & \tablehdg{$\mu$} & \tablehdg{$\sigma$} & \tablehdg{$\mu$} & \tablehdg{$\sigma$} & \tablehdg{$\mu$} & \tablehdg{$\sigma$} & \tablehdg{$\mu$} & \tablehdg{$\sigma$} & \tablehdg{$\mu$} & \tablehdg{$\sigma$} \\
          \midrule
           \textit{Hover} & \textbf{.022} & \textbf{.014} & .058 & .067 & \textbf{.020} & \textbf{.014} & .062 & .049 & \textbf{.036} & \textbf{.039} & .087 & .050\\
           \textit{Forward} & \textbf{.015} & \textbf{.012} & .059 & .066 & \textbf{.012} & \textbf{.010} & .121 & .065 & \textbf{.013} & \textbf{.011} & .017 & .027\\
           \textit{Lateral} & \textbf{.022} & \textbf{.018} & .147 & .076 & \textbf{.022} & \textbf{.017} & .055 & .073 & \textbf{.019} & \textbf{.016} & .062 & .047\\
           \textit{Vertical} & \textbf{.018} & \textbf{.014} & .049 & .054 & \textbf{.018} & \textbf{.015} & .214 & .068 & \textbf{.018} & \textbf{.016} & .036 & .022\\
           \textit{Rectangle} & \textbf{.021} & \textbf{.017} & .178 & .053 & \textbf{.021} & \textbf{.014} & .130 & .062 & \textbf{.030} & \textbf{.021} & .040 & .045\\
           \textit{Loop} & \textbf{.029} & \textbf{.021} & .148 & .084 & \textbf{.026} & \textbf{.020} & .173 & .093 & \textbf{.052} & \textbf{.043} & .082 & .066\\
          \bottomrule
        \end{tabular}
    \end{table}

In summary, \ac{maslo} is better at estimating the linear motion, while \ac{liosam} estimated the rotational motion more accurately.
The complementary strengths of the evaluated algorithms could be combined into one algorithm, with both \ac{mas} and inertial factors.
We set this task as future work.

\section{Conclusion and Future Work}
\label{section:conclusion}

We have introduced our approach of \acf{mas} preintegration to be used as a more accurate alternative for the \ac{imu} preintegration used in most existing \ac{uav} state estimation algorithms.
By directly measuring the \ac{mas}, we eliminated the need for using \ac{imu}, therefore with the proposed method it is not necessary to design hardware solutions for damping the measurement noise caused by the vibrations of spinning propellers.
Further, we have presented the \ac{maslo} algorithm, which combines the preintegrated \ac{mas} measurements in the \ac{mas} factor with \ac{lidar} odometry to provide a precise estimation of the whole \ac{uav} state.
We have shown improved estimation of the translational part of the \ac{uav} state w.r.t. inertial preintegration on outdoor flight datasets with accurate \ac{rtk} ground truth.
\ac{maslo} outperformed \ac{liosam} by 28\% in the \ac{ate} metric and 65\% in the \ac{ave} metric.
Despite our state estimation method being model-based, with parameters differing for every \ac{uav} platform, we have shown its robustness to inaccurately set parameter values.
For most parameters, setting the value even orders of magnitude wrong still allows the algorithm to perform reasonably well.

Based on the results, we see a future research direction in combining our \ac{mas} factor with the \ac{imu} factor in a single state estimation algorithm that would exploit the complementary strengths of both factors.
Another research opportunity is to introduce our \ac{mas} factor into visual localization algorithms, where it can replace or supplement the role of the \ac{imu} factor similarly as in \ac{maslo}.
\ac{vio} methods are very sensitive to the quality of \ac{imu} data, because the inertial measurements are necessary to recover the metric scale of the motion.
Using the proposed \ac{mas} measurements would eliminate the need for calibration of \ac{imu}, designing hardware damping, and software filtering.
The \ac{lidar} odometry part of \ac{liosam}, which we base \ac{maslo} on, was also recently outperformed by more recent \ac{ekf}-based approaches, such as Fast-LIO2 \cite{xu2022fast} or Point-LIO \cite{he2023point}, which could also benefit from incorporating \ac{mas} preintegration into the \ac{ekf} formulation.
Finally, the dynamic model used in this paper is simplified by neglecting some forces and torques exerted by the propellers, which are assumed to be sufficiently small and compensated for by the noise model.
However, including these forces and torques in a more realistic model has the potential to improve the quality of the state estimate.

By developing our work using the widely used GTSAM framework, we believe it can be easily incorporated into new state estimation algorithms for \acp{uav} to improve their performance.
This is also the reason for open-sourcing both the developed algorithms and the recorded datasets in order to encourage future research and development in the \ac{mas}-based state estimation.



\bibliographystyle{elsarticle-num}
\bibliography{main.bib}

@article{gonzalez2013real,
  title={{Real-time attitude stabilization of a mini-uav quad-rotor using motor speed feedback}},
  author={Gonz{\'a}lez, Iv{\'a}n and Salazar, Sergio and Torres, Jorge and Lozano, Rogelio and Romero, Hugo},
  journal={Journal of Intelligent \& Robotic Systems},
  volume={70},
  pages={93--106},
  year={2013},
  publisher={Springer}
}

@article{cioffi2023learned,
  title = {Learned {{Inertial Odometry}} for {{Autonomous Drone Racing}}},
  author = {Cioffi, Giovanni and Bauersfeld, Leonard and Kaufmann, Elia and Scaramuzza, Davide},
  year = {2023},
  month = may,
  journal = {IEEE Robotics and Automation Letters},
  volume = {8},
  number = {5},
  pages = {2684--2691},
}

@inproceedings{antonini2018blackbird,
  title={{The Blackbird Dataset: A large-scale dataset for UAV perception in aggressive flight}},
  booktitle={2018 International Symposium on Experimental Robotics (ISER)},
  author={
    Antonini, Amado and 
    Guerra, Winter and 
    Murali, Varun and 
    Sayre-McCord, Thomas and 
    Karaman, Sertac},
  year={2018}
}

@INPROCEEDINGS{burri2015robust,
  author={Burri, Michael and Dätwiler, Manuel and Achtelik, Markus W. and Siegwart, Roland},
  booktitle={2015 IEEE International Conference on Robotics and Automation (ICRA)},
  title={{Robust state estimation for Micro Aerial Vehicles based on system dynamics}},
  year={2015},
  volume={},
  number={},
  pages={5278-5283}
}

@article{sahili2023SurveyVisual,
  title = {A {{Survey}} of {{Visual SLAM Methods}}},
  author = {Sahili, Ali Rida and Hassan, Saifeldin and Sakhrieh, Saber Muawiyah and Mounsef, Jinane and Maalouf, Noel and Arain, Bilal and Taha, Tarek},
  year = {2023},
  journal = {IEEE Access},
  volume = {11},
  pages = {139643--139677},
  issn = {2169-3536},
  urldate = {2024-01-30},
}

@article{lee2024lidar,
  title={LiDAR odometry survey: recent advancements and remaining challenges},
  author={Lee, Dongjae and Jung, Minwoo and Yang, Wooseong and Kim, Ayoung},
  journal={Intelligent Service Robotics},
  volume={17},
  number={2},
  pages={95--118},
  year={2024},
  publisher={Springer}
}

@inproceedings{qin2020lins,
  title={{Lins: A lidar-inertial state estimator for robust and efficient navigation}},
  author={Qin, Chao and Ye, Haoyang and Pranata, Christian E and Han, Jun and Zhang, Shuyang and Liu, Ming},
  booktitle={2020 IEEE international conference on robotics and automation (ICRA)},
  pages={8899--8906},
  year={2020},
  organization={IEEE}
}

@article{xu2021fast,
  title={{Fast-lio: A fast, robust lidar-inertial odometry package by tightly-coupled iterated kalman filter}},
  author={Xu, Wei and Zhang, Fu},
  journal={IEEE Robotics and Automation Letters},
  volume={6},
  number={2},
  pages={3317--3324},
  year={2021},
  publisher={IEEE}
}

@article{xu2022fast,
  title={{Fast-lio2: Fast direct lidar-inertial odometry}},
  author={Xu, Wei and Cai, Yixi and He, Dongjiao and Lin, Jiarong and Zhang, Fu},
  journal={IEEE Transactions on Robotics},
  volume={38},
  number={4},
  pages={2053--2073},
  year={2022},
  publisher={IEEE}
}

@article{forster2016manifold,
  title={{On-manifold preintegration for real-time visual--inertial odometry}},
  author={Forster, Christian and Carlone, Luca and Dellaert, Frank and Scaramuzza, Davide},
  journal={IEEE Transactions on Robotics},
  volume={33},
  number={1},
  pages={1--21},
  year={2016},
  publisher={IEEE}
}

@inproceedings{geneva2018lips,
  title={{Lips: Lidar-inertial 3d plane slam}},
  author={Geneva, Patrick and Eckenhoff, Kevin and Yang, Yulin and Huang, Guoquan},
  booktitle={2018 IEEE/RSJ International Conference on Intelligent Robots and Systems (IROS)},
  pages={123--130},
  year={2018},
  organization={IEEE}
}

@inproceedings{shan2020lio,
  title={{Lio-sam: Tightly-coupled lidar inertial odometry via smoothing and mapping}},
  author={Shan, Tixiao and Englot, Brendan and Meyers, Drew and Wang, Wei and Ratti, Carlo and Rus, Daniela},
  booktitle={2020 IEEE/RSJ international conference on intelligent robots and systems (IROS)},
  pages={5135--5142},
  year={2020},
  organization={IEEE}
}

@electronic{dellaert2022gtsam,
  author       = {Frank Dellaert and GTSAM Contributors},
  title        = {borglab/gtsam},
  month        = May,
  year         = 2022,
  publisher    = {Georgia Tech Borg Lab},
  version      = {4.2a8},
  url          = {https://github.com/borglab/gtsam)}}

@manual{dellaert2012factor,
  title={{Factor graphs and GTSAM: A hands-on introduction}},
  author={Dellaert, Frank},
  journal={Georgia Institute of Technology, Tech. Rep},
  volume={2},
  pages={4},
  year={2012}
}

@inproceedings{juric2021ComparisonGraph,
  title = {A {{Comparison}} of {{Graph Optimization Approaches}} for {{Pose Estimation}} in {{SLAM}}},
  booktitle = {2021 44th {{International Convention}} on {{Information}}, {{Communication}} and {{Electronic Technology}} ({{MIPRO}})},
  author = {Juric, Andela and Kendes, Filip and Markovic, Ivan and Petrovic, Ivan},
  year = {2021},
  month = sep,
  pages = {1113--1118},
  publisher = {{IEEE}},
  address = {{Opatija, Croatia}},
  urldate = {2024-01-31},
  isbn = {978-953-233-101-1},
}

@article{kratky2021exploration,
	author = "Krátký, Vít and Petráček, Pavel and Báča, Tomáš and Saska, Martin",
	journal = "Journal of Field Robotics",
	month = "May",
	number = 8,
	pages = "1036-1058",
	title = "An autonomous unmanned aerial vehicle system for fast exploration of large complex indoor environments",
	volume = 38,
	year = 2021
}

@article{walter2022fr,
	author = "Walter, V. and Spurny, V. and Petrlik, M. and Baca, T. and Zaitlík, D. and Demkiv, L. and and Saska, M.",
	issn = "2771-3989",
	journal = "Field Robotics",
	month = "April",
	pages = "406--436",
	title = "{Extinguishing real fires by fully autonomous multirotor UAVs in the MBZIRC 2020 competition}",
	volume = 2,
	year = 2022
}

@article{stibinger2020ral,
	author = "P. {\v{S}tibinger} and T. {B\'{a}\v{c}a} and M. {Saska}",
	issn = "2377-3766",
	journal = "IEEE Robotics and Automation Letters",
	month = "April",
	number = 2,
	pages = "3634-3641",
	title = "Localization of Ionizing Radiation Sources by Cooperating Micro Aerial Vehicles With Pixel Detectors in Real-Time",
	volume = 5,
	year = 2020
}

@article{petrlik2020robust,
	author = "M. {Petrl\'{i}k} and T. {B\'{a}\v{c}a} and D. {He\v{r}t} and M. {Vrba} and T. {Krajn\'{i}k} and M. {Saska}",
	issn = "2377-3766",
	journal = "IEEE Robotics and Automation Letters",
	month = "April",
	number = 2,
	pages = "2169-2176",
	title = "A Robust UAV System for Operations in a Constrained Environment",
	volume = 5,
	year = 2020
}

@article{petracek2021caves,
	author = "Petracek, Pavel and Kratky, Vit and Petrlik, Matej and Baca, Tomas and Kratochvil, Radim and Saska, Martin",
	journal = "IEEE Robotics and Automation Letters",
	month = "October",
	number = 4,
	pages = "7596-7603",
	title = "Large-Scale Exploration of Cave Environments by Unmanned Aerial Vehicles",
	volume = 6,
	year = 2021
}

@article{orekhov2023inspiring,
	author = "Orekhov, V. and Maio, A. and Daniel, R. and Chung, T.",
	issn = "2771-3989",
	journal = "Field Robotics",
	month = "April",
	pages = "560--604",
	title = "{Inspiring Field Robotics Advances through the Design of the DARPA Subterranean Challenge}",
	volume = 3,
	year = 2023
}

@article{zhang2017low,
  title={{Low-drift and real-time lidar odometry and mapping}},
  author={Zhang, Ji and Singh, Sanjiv},
  journal={Autonomous Robots},
  volume={41},
  pages={401--416},
  year={2017},
  publisher={Springer}
}

@article{romero2022time,
  title={{Time-optimal online replanning for agile quadrotor flight}},
  author={Romero, Angel and Penicka, Robert and Scaramuzza, Davide},
  journal={IEEE Robotics and Automation Letters},
  volume={7},
  number={3},
  pages={7730--7737},
  year={2022},
  publisher={IEEE}
}

@article{foehn2022agilicious,
  title={{Agilicious: Open-source and open-hardware agile quadrotor for vision-based flight}},
  author={Foehn, Philipp and Kaufmann, Elia and Romero, Angel and Penicka, Robert and Sun, Sihao and Bauersfeld, Leonard and Laengle, Thomas and Cioffi, Giovanni and Song, Yunlong and Loquercio, Antonio and others},
  journal={Science robotics},
  volume={7},
  number={67},
  pages={eabl6259},
  year={2022},
  publisher={American Association for the Advancement of Science}
}

@article{bosse2012zebedee,
  title={{Zebedee: Design of a spring-mounted 3-d range sensor with application to mobile mapping}},
  author={Bosse, Michael and Zlot, Robert and Flick, Paul},
  journal={IEEE Transactions on Robotics},
  volume={28},
  number={5},
  pages={1104--1119},
  year={2012},
  publisher={IEEE}
}

@inproceedings{kummerle2011g2o,
  title={{g 2 o: A general framework for graph optimization}},
  author={K{\"u}mmerle, Rainer and Grisetti, Giorgio and Strasdat, Hauke and Konolige, Kurt and Burgard, Wolfram},
  booktitle={2011 IEEE International Conference on Robotics and Automation},
  pages={3607--3613},
  year={2011},
  organization={IEEE}
}

@article{mur2015orb,
  title={{ORB-SLAM: a versatile and accurate monocular SLAM system}},
  author={Mur-Artal, Raul and Montiel, Jose Maria Martinez and Tardos, Juan D},
  journal={IEEE transactions on robotics},
  volume={31},
  number={5},
  pages={1147--1163},
  year={2015},
  publisher={IEEE}
}

@article{mur2017orb,
  title={{Orb-slam2: An open-source slam system for monocular, stereo, and rgb-d cameras}},
  author={Mur-Artal, Raul and Tard{\'o}s, Juan D},
  journal={IEEE transactions on robotics},
  volume={33},
  number={5},
  pages={1255--1262},
  year={2017},
  publisher={IEEE}
}

@article{forster2016svo,
  title={{SVO: Semidirect visual odometry for monocular and multicamera systems}},
  author={Forster, Christian and Zhang, Zichao and Gassner, Michael and Werlberger, Manuel and Scaramuzza, Davide},
  journal={IEEE Transactions on Robotics},
  volume={33},
  number={2},
  pages={249--265},
  year={2016},
  publisher={IEEE}
}

@article{agarwal2012ceres,
  title={{Ceres solver: Tutorial \& reference}},
  author={Agarwal, Sameer and Mierle, Keir},
  journal={Google Inc},
  volume={2},
  number={72},
  pages={8},
  year={2012}
}

@article{leutenegger2015keyframe,
  title={{Keyframe-based visual--inertial odometry using nonlinear optimization}},
  author={Leutenegger, Stefan and Lynen, Simon and Bosse, Michael and Siegwart, Roland and Furgale, Paul},
  journal={The International Journal of Robotics Research},
  volume={34},
  number={3},
  pages={314--334},
  year={2015},
  publisher={SAGE Publications Sage UK: London, England}
}

@article{qin2018vins,
  title={{Vins-mono: A robust and versatile monocular visual-inertial state estimator}},
  author={Qin, Tong and Li, Peiliang and Shen, Shaojie},
  journal={IEEE Transactions on Robotics},
  volume={34},
  number={4},
  pages={1004--1020},
  year={2018},
  publisher={IEEE}
}

@inproceedings{geneva2020openvins,
  title={{Openvins: A research platform for visual-inertial estimation}},
  author={Geneva, Patrick and Eckenhoff, Kevin and Lee, Woosik and Yang, Yulin and Huang, Guoquan},
  booktitle={2020 IEEE International Conference on Robotics and Automation (ICRA)},
  pages={4666--4672},
  year={2020},
  organization={IEEE}
}

@inproceedings{rosinol2020kimera,
  title={{Kimera: an open-source library for real-time metric-semantic localization and mapping}},
  author={Rosinol, Antoni and Abate, Marcus and Chang, Yun and Carlone, Luca},
  booktitle={2020 IEEE International Conference on Robotics and Automation (ICRA)},
  pages={1689--1696},
  year={2020},
  organization={IEEE}
}

@inproceedings{furgale2013unified,
  title={{Unified temporal and spatial calibration for multi-sensor systems}},
  author={Furgale, Paul and Rehder, Joern and Siegwart, Roland},
  booktitle={2013 IEEE/RSJ International Conference on Intelligent Robots and Systems},
  pages={1280--1286},
  year={2013},
  organization={IEEE}
}

@inproceedings{qin2018online,
  title={{Online temporal calibration for monocular visual-inertial systems}},
  author={Qin, Tong and Shen, Shaojie},
  booktitle={2018 IEEE/RSJ International Conference on Intelligent Robots and Systems (IROS)},
  pages={3662--3669},
  year={2018},
  organization={IEEE}
}

@inproceedings{yang2020OnlineIMU,
  title = {Online {{IMU Intrinsic Calibration}}: {{Is It Necessary}}?},
  shorttitle = {Online {{IMU Intrinsic Calibration}}},
  booktitle = {Robotics: {{Science}} and {{Systems XVI}}},
  author = {Yang, Yulin and Geneva, Patrick and Zuo, Xingxing and Huang, Guoquan},
  year = {2020},
  month = jul,
  publisher = {{Robotics: Science and Systems Foundation}},
  urldate = {2024-02-02},
  isbn = {978-0-9923747-6-1},
  langid = {english},
}

@article{niu2015quantitative,
  title={{Quantitative analysis to the impacts of IMU quality in GPS/INS deep integration}},
  author={Niu, Xiaoji and Ban, Yalong and Zhang, Quan and Zhang, Tisheng and Zhang, Hongping and Liu, Jingnan},
  journal={Micromachines},
  volume={6},
  number={8},
  pages={1082--1099},
  year={2015},
  publisher={MDPI}
}

@inproceedings{duan2020dynamical,
  title={{Dynamical Analysis for the INS Vibration Control System Used in UAV}},
  author={Duan, Yuxing and Li, Xiao and Su, Bo and Wang, Xin and Yang, Qiang},
  booktitle={IOP Conference Series: Materials Science and Engineering},
  volume={887},
  number={1},
  pages={012026},
  year={2020},
  organization={IOP Publishing}
}

@inproceedings{li2017development,
  title={{Development and design methodology of an anti-vibration system on micro-UAVs}},
  author={Li, Zhenming and Lao, Mingjie and Phang, Swee King and Hamid, Mohamed Redhwan Abdul and Tang, Kok Zuea and Lin, Feng},
  booktitle={International micro air vehicle conference and flight competition (IMAV)},
  pages={223--228},
  year={2017}
}

@article{capriglione2020experimental,
  title={{Experimental analysis of filtering algorithms for IMU-based applications under vibrations}},
  author={Capriglione, Domenico and Carrat{\`u}, Marco and Catelani, Marcantonio and Ciani, Lorenzo and Patrizi, Gabriele and Pietrosanto, Antonio and Sommella, Paolo},
  journal={IEEE Transactions on Instrumentation and Measurement},
  volume={70},
  pages={1--10},
  year={2020},
  publisher={IEEE}
}

@inproceedings{bouabdallah2007full,
  title={{Full control of a quadrotor}},
  author={Bouabdallah, Samir and Siegwart, Roland},
  booktitle={2007 IEEE/RSJ international conference on intelligent robots and systems},
  pages={153--158},
  year={2007},
  organization={Ieee}
}

@article{lupton2011visual,
  title={{Visual-inertial-aided navigation for high-dynamic motion in built environments without initial conditions}},
  author={Lupton, Todd and Sukkarieh, Salah},
  journal={IEEE Transactions on Robotics},
  volume={28},
  number={1},
  pages={61--76},
  year={2011},
  publisher={IEEE}
}

@article{sola2018micro,
  title={{A micro Lie theory for state estimation in robotics}},
  author={Sola, Joan and Deray, Jeremie and Atchuthan, Dinesh},
  journal={arXiv preprint arXiv:1812.01537},
  year={2018}
}

@article{hert2023mrs,
  title={{MRS drone: A modular platform for real-world deployment of aerial multi-robot systems}},
  author={Hert, Daniel and Baca, Tomas and Petracek, Pavel and Kratky, Vit and Penicka, Robert and Spurny, Vojtech and Petrlik, Matej and Vrba, Matous and Zaitlik, David and Stoudek, Pavel and others},
  journal={Journal of Intelligent \& Robotic Systems},
  volume={108},
  number={4},
  pages={64},
  year={2023},
  publisher={Springer}
}

@electronic{grupp2017evo,
  title={{evo: Python package for the evaluation of odometry and SLAM.}},
  author={Grupp, Michael},
  howpublished={\url{https://github.com/MichaelGrupp/evo}},
  year={2017}
}

@inproceedings{zhang2018tutorial,
  title={{A tutorial on quantitative trajectory evaluation for visual (-inertial) odometry}},
  author={Zhang, Zichao and Scaramuzza, Davide},
  booktitle={2018 IEEE/RSJ International Conference on Intelligent Robots and Systems (IROS)},
  pages={7244--7251},
  year={2018},
  organization={IEEE}
}

@article{he2023point,
  title={{Point-LIO: Robust High-Bandwidth Light Detection and Ranging Inertial Odometry}},
  author={He, Dongjiao and Xu, Wei and Chen, Nan and Kong, Fanze and Yuan, Chongjian and Zhang, Fu},
  journal={Advanced Intelligent Systems},
  volume={5},
  number={7},
  pages={2200459},
  year={2023},
  publisher={Wiley Online Library}
}

@article{petracek2024rms,
	author = "Petracek, Pavel and Alexis, Kostas and Saska, Martin",
	journal = "IEEE Robotics and Automation Letters",
	number = 6,
	pages = "5230--5237",
	title = "{RMS: Redundancy-Minimizing Point Cloud Sampling for Real-Time Pose Estimation}",
	volume = 9,
	year = 2024
}

@article{kaess2012isam2,
  title={{iSAM2: Incremental smoothing and mapping using the Bayes tree}},
  author={Kaess, Michael and Johannsson, Hordur and Roberts, Richard and Ila, Viorela and Leonard, John J and Dellaert, Frank},
  journal={The International Journal of Robotics Research},
  volume={31},
  number={2},
  pages={216--235},
  year={2012},
  publisher={Sage Publications Sage UK: London, England}
}

@inproceedings{kaess2011bayes,
  title={{The Bayes tree: An algorithmic foundation for probabilistic robot mapping}},
  author={Kaess, Michael and Ila, Viorela and Roberts, Richard and Dellaert, Frank},
  booktitle={Algorithmic Foundations of Robotics IX: Selected Contributions of the Ninth International Workshop on the Algorithmic Foundations of Robotics},
  pages={157--173},
  year={2011},
  organization={Springer}
}

@inproceedings{meier2015px4,
  title={{PX4: A node-based multithreaded open source robotics framework for deeply embedded platforms}},
  author={Meier, Lorenz and Honegger, Dominik and Pollefeys, Marc},
  booktitle={2015 IEEE international conference on robotics and automation (ICRA)},
  pages={6235--6240},
  year={2015},
  organization={IEEE}
}

@article{ebadi2023present,
  title={{Present and future of slam in extreme environments: The darpa subt challenge}},
  author={Ebadi, Kamak and Bernreiter, Lukas and Biggie, Harel and Catt, Gavin and Chang, Yun and Chatterjee, Arghya and Denniston, Christopher E and Desch{\^e}nes, Simon-Pierre and Harlow, Kyle and Khattak, Shehryar and others},
  journal={IEEE Transactions on Robotics},
  year={2023},
  publisher={IEEE}
}

@article{baca2021mrs,
  title={{The MRS UAV system: Pushing the frontiers of reproducible research, real-world deployment, and education with autonomous unmanned aerial vehicles}},
  author={Baca, Tomas and Petrlik, Matej and Vrba, Matous and Spurny, Vojtech and Penicka, Robert and Hert, Daniel and Saska, Martin},
  journal={Journal of Intelligent \& Robotic Systems},
  volume={102},
  number={1},
  pages={26},
  year={2021},
  publisher={Springer}
}

@article{cioffi2023hdvio,
  title={{Hdvio: Improving localization and disturbance estimation with hybrid dynamics vio}},
  author={Cioffi, Giovanni and Bauersfeld, Leonard and Scaramuzza, Davide},
  journal={arXiv preprint arXiv:2306.11429},
  year={2023}
}

@inproceedings{zhang2022visual,
  title={{The visual-inertial-dynamical multirotor dataset}},
  author={Zhang, Kunyi and Yang, Tiankai and Ding, Ziming and Yang, Sheng and Ma, Teng and Li, Mingyang and Xu, Chao and Gao, Fei},
  booktitle={2022 International Conference on Robotics and Automation (ICRA)},
  pages={7635--7641},
  year={2022},
  organization={IEEE}
}

@article{nisar2019vimo,
  title={{Vimo: Simultaneous visual inertial model-based odometry and force estimation}},
  author={Nisar, Barza and Foehn, Philipp and Falanga, Davide and Scaramuzza, Davide},
  journal={IEEE Robotics and Automation Letters},
  volume={4},
  number={3},
  pages={2785--2792},
  year={2019},
  publisher={IEEE}
}

@article{svacha2020imu,
  title={{Imu-based inertia estimation for a quadrotor using newton-euler dynamics}},
  author={Svacha, James and Paulos, James and Loianno, Giuseppe and Kumar, Vijay},
  journal={IEEE Robotics and Automation Letters},
  volume={5},
  number={3},
  pages={3861--3867},
  year={2020},
  publisher={IEEE}
}

@article{penicka2022minimum,
    title={{Minimum-time quadrotor waypoint flight in cluttered environments}},
    author={Penicka, Robert and Scaramuzza, Davide},
    journal={IEEE Robotics and Automation Letters},
    volume={7},
    number={2},
    pages={5719--5726},
    year={2022},
    publisher={IEEE}
}

@article{brossard2020denoising,
    title={{Denoising imu gyroscopes with deep learning for open-loop attitude estimation}},
    author={Brossard, Martin and Bonnabel, Silvere and Barrau, Axel},
    journal={IEEE Robotics and Automation Letters},
    volume={5},
    number={3},
    pages={4796--4803},
    year={2020},
    publisher={IEEE}
}

@article{tal2020accurate,
    title={Accurate tracking of aggressive quadrotor trajectories using incremental nonlinear dynamic inversion and differential flatness},
    author={Tal, Ezra and Karaman, Sertac},
    journal={IEEE Transactions on Control Systems Technology},
    volume={29},
    number={3},
    pages={1203--1218},
    year={2020},
    publisher={IEEE}
}

@article{sieberling2010robust,
    title={Robust flight control using incremental nonlinear dynamic inversion and angular acceleration prediction},
    author={Sieberling, S and Chu, QP and Mulder, JA},
    journal={Journal of guidance, control, and dynamics},
    volume={33},
    number={6},
    pages={1732--1742},
    year={2010}
}

@inproceedings{more2006levenberg,
  title={The Levenberg-Marquardt algorithm: implementation and theory},
  author={More, Jorge J},
  booktitle={Numerical analysis: proceedings of the biennial Conference held at Dundee, June 28--July 1, 1977},
  pages={105--116},
  year={2006},
  organization={Springer}
}

\end{document}